\documentclass[manuscript,screen]{acmart}
\AtBeginDocument{%
  \providecommand\BibTeX{{%
    \normalfont B\kern-0.5em{\scshape i\kern-0.25em b}\kern-0.8em\TeX}}}

\usepackage{booktabs}
\usepackage{lscape}
\usepackage[linguistics]{forest,adjustbox}
\usepackage{graphicx}
\usepackage{subfigure}

\newcommand\tab[1][1cm]{\hspace*{#1}}

\begin{document}


\title[Resources, Statistics and Applications for Three Purvanchal Languages]{Linguistic Resources for Bhojpuri, Magahi and Maithili: Statistics about them, their Similarity Estimates, and Baselines for Three Applications}

\author{Rajesh Kumar Mundotiya}
\email{rajeshkm.rs.cse16@iitbhu.ac.in}
\affiliation{%
\institution{Indian Institute of Technology (BHU), India}
  \city{Varanasi}
  \state{U.P.}
}

\author{Manish Kumar Singh}
\email{maneeshhsingh100@gmail.com}
\affiliation{%
  \institution{Alexa AI, Amazon, India}
}

\author{Rahul Kapur}
\email{rahul18182212@gmail.com}
\affiliation{%
  \institution{Samsung Research Institute, India}
  \city{Noida}
  \state{U.P.}
}

\author{Swasti Mishra}
\email{swasti.hss@iitbhu.ac.in}
\affiliation{%
  \institution{Indian Institute of Technology (BHU), India}
  \city{Varanasi}
  \state{U.P.}
}

\author{Anil Kumar Singh}
\email{aksingh.cse@iitbhu.ac.in}
\affiliation{%
  \institution{Indian Institute of Technology (BHU), India}
  \city{Varanasi}
  \state{U.P.}
}

\renewcommand{\shortauthors}{Mundotiya et al.}

\begin{abstract}
 Corpus preparation for low-resource languages and for development of human language technology to analyze or computationally process them is a laborious task, primarily due to the unavailability of expert linguists who are native speakers of these languages and also due to the time and resources required. Bhojpuri, Magahi, and Maithili, languages of the Purvanchal region of India (in the north-eastern parts), are low-resource languages belonging to the Indo-Aryan (or Indic) family. They are closely related to Hindi, which is a relatively high-resource language, which is why we compare with Hindi. We collected corpora for these three languages from various sources and cleaned them to the extent possible, without changing the data in them. The text belongs to different domains and genres. We calculated some basic statistical measures for these corpora at character, word, syllable, and morpheme levels. These corpora were also annotated with parts-of-speech (POS) and chunk tags. The basic statistical measures were both absolute and relative and were exptected to indicate of linguistic properties such as morphological, lexical, phonological, and syntactic complexities (or richness). The results were compared with a standard Hindi corpus. For most of the measures, we tried to the corpus size the same across the languages to avoid the effect of corpus size, but in some cases it turned out that using the full corpus was better, even if sizes were very different. Although the results are not very clear, we try to draw some conclusions about the languages and the corpora. For POS tagging and chunking, the BIS tagset was used to manually annotate the data. The POS tagged data sizes are 16067, 14669 and 12310 sentences, respectively, for Bhojpuri, Magahi and Maithili. The sizes for chunking are 9695 and 1954 sentences for Bhojpuri and Maithili, respectively. The inter-annotator agreement for these annotations, using Cohen's Kappa, was 0.92, 0.64, and 0.74, respectively, for the three languages. These (annotated) corpora have been used for developing preliminary automated tools, which include POS tagger, Chunker and Language Identifier. We have also developed the Bilingual dictionary (Purvanchal languages to Hindi) and a Synset (that can be integrated later in the Indo-WordNet) as additional resources. The main contribution of the work is the creation of basic resources for facilitating further language processing research for these languages, providing some quantitative measures about them and their similarities among themselves and with Hindi. For similarities, we use a somewhat novel measure of language similarity based on an n-gram based language identification algorithm. An additional contribution is providing baselines for three basic NLP applications (POS tagging, chunking and language identification) for these closely related languages.
\end{abstract}

\ccsdesc[500]{Computing methodologies~Language resources}
\ccsdesc[100]{Computing methodologies~Machine translation}
\ccsdesc[300]{Computing methodologies~Lexical semantics}
\ccsdesc[300]{Computing methodologies~Phonology / morphology}

\keywords{Corpus, Syntactic annotation, Low resource language, Inter-annotator agreement, POS tagging, Chunking, Language Identification, Language similarity}

\maketitle

\section{Introduction}
Plain text corpus, from the point of view of Computational Linguistics (CL) or Natural Language Processing (NLP), is a systematic collection of text that should have certain properties like representativeness, correctness (if it is for standard written language) and sufficient quantity. For languages on which no NLP work has been performed before, or very little such work exists, creating a good plain text corpus is the foundation for building NLP resources and applications or tools for that language. Plain text corpus is basically a collection of written text or transcribed speeches of a particular natural language stored in electronic or digital form. Linguists and NLP researchers use corpora for statistical analysis, hypothesis testing, word-frequency analysis and validating linguistic knowledge in a language territory or domain. Once this corpus gets annotation with richer linguistic knowledge about its structure, meaning or usage etc., it can act as training data for supervised learning for building tools that solve various NLP problems such as Part-of-Speech (POS) tagging, chunking, morphological analysis, parsing etc. In this paper, we describe our efforts to prepare some plain text corpus for three languages of the Purvanchal region of central north-eastern parts of India and to annotate this corpus with POS and chunk tags. These languages are Bhojpuri, Maithili, and Magahi.

These languages are low-resource languages and are phylogenetically related since they belong to the same (Indo-Aryan, or Indic) language family and also mainly occur in a geographically contiguous area. Globally, these are the first-languages of 50, 12, and 30 million speakers\footnote{\url{https://www.censusindia.gov.in/2011Census/Language-2011/Statement-1.pdf}}, respectively. The availability of an enormous base of speakers is a good reason for researchers to understand and analyze different properties of these languages and to build language technologies for them. \citet{kumar2011challenges, singh2015statistical} began a project by collecting corpora of 1 million words for the Bhojpuri and Magahi languages, out of which 89997 and 63000 words have been annotated at the POS level, respectively. Simultaneously, at our end, we have collected corpora for the Bhojpuri, Magahi, and Maithili, consisting of 704543, 424434, and 267881 words. We have also carried out manual POS annotation on corpus sizes of 245482, 171509 and 208640 words, respectively, and the chunk-annotated corpus sizes of 60588, 0 and 10476 words, respectively, as chunk annotation for Maithili could not be carried out. 

To the best of our knowledge, our effort, which began in 2014, was one of the first to create annotated resources for these languages. At that time, there were hardly any publicly available sources of Maithili or Bhojpuri text corpus, and it was the first attempt to build the corpora and annotate them for POS tags and chunks. The project also involved building language processing tools for these languages and, in the end, to build machine translations systems for these languages. 

These linguistically annotated basic resources can serve as the base for further developing language technology and tools for these languages such as POS tagger, Chunker, Morphological Analyzer, Morphological Generator, Machine Translation, Cognate Identification etc. can be carried out by researchers. This paper describes these basic resources for three Purvanchal languages and some basic statistical details about them at the word, the morpheme, the syllable, and the character levels, and their similarities among themselves and with Hindi. These details include some simple measures of complexity and similarity, including a slighthly novel similarity measure. Apart from this, we also highlight some issues encountered during the Part-of-Speech and Chunking annotations for all these languages. We have also tried to pay some special attention to the similarities among these languages, as their differences are still debated. The inter-annotator agreement has been calculated through Cohen's Kappa (\citet{doi:10.1177/001316446002000104}). The (annotated) corpora have been employed to build the POS taggers, Chunkers and a Language identifier for the three languages.

\section{Related Work}
India, as is well known, is a multilingual country: it has 22 scheduled (constitutionally recognised) languages, written in (at least) 13 different extant scripts and several other variations of them. Apart from the scheduled languages, there are at least 400 plus other languages.  Most of the developed countries, especially the ones with a not so diverse language base compared to India, do possess relatively more abundant, easily accessible, and most importantly, good quality corpora. On the other hand, Indian languages have relatively little corpora for a large number of languages, even though the degree of resource-scarceness varies. This scarcity severely constraints the language technology development and application of state-of-the-art algorithms, particularly deep learning approaches.

Technology Development for Indian Languages or TDIL\footnote{\url{http://www.tdil.meity.gov.in}} (\citet{dash2004language}), a Government of India department under the Ministry of Electronic and Information Technology, has been actively promoting work on language resource and technology development for Indian languages, particularly on the scheduled languages. As an example of work done as part of this initiative, machine-readable corpora of around 10M tokens of Indian languages was prepared. The languages include English, Hindi, Punjabi, Telugu, Kannada, Tamil, Malayalam, Marathi, Gujarati, Oriya, Bangla, Assamese, Urdu, Sindhi, Kashmiri, and Sanskrit. Out of these, Urdu, Kashmiri, and Sindhi use Perso-Arabic scripts for their writing systems, while the remaining languages use scripts derived from the ancient Brahmi script. All these languages belong to either the Dravidian and or the Indo-Aryan families. The preliminary analysis of some of their initial corpora is described in \citet{bharati2000basic}. The resources and tools or technologies developed for Indian languages and shared with the TDIL are distributed through the Indian Languages Linguistic Data Consortium (IL-LDC)\footnote{\url{http://www.ildc.in/}}.

Effort was made by various research centres working with TDIL to create representative corpora for Indian languages and basic analysis of such corpora was published. For example, the Bangla corpus, that was compiled from various disciplines such as literature, mass communication, science, etc. and was initially subjected to statistical analysis by Chaudhuri and Ghosh (\citet{chaudhuri1998statistical}) at word, character, and sentence levels on the monolingual corpus comprising 2.5 million words from the Anusaaraka group\footnote{\url{https://ltrc.iiit.ac.in/projects/Anusaaraka/anu_home.html}} and 150,000 words from their own laboratory. A further collection of the corpus from the Bangla newspaper website (Prothom-Alo) was crawled, considering the balance and representativeness of the corpus. Its comparison with the earlier corpus (\citet{majumder2006analysis}) was also reported. In addition to this, to boost the language technology development for Bangla, Bangla-translated Basic Travel Expression Corpus (BTEC) (\citet{khan2014bengali}) was compiled to develop an English-Bangla speech translation system in the travel domain.

Another corpora analysis under the TDIL program was for the Telugu language, which is a morphologically rich language, much more so than Bhojpuri, Maithili, and Magahi. It is also one of the most spoken languages of India (ranking third in terms of its speakers), and mainly based geographically in the southern states of Andhra Pradesh and Telangana, India. Its characteristics were analyzed using statistical tools on the compiled monolingual corpus of Telugu comprising 39 million words (\citet{kumar2007statistical}). Moreover, lexical categories produced from the morphological analyzer were used as parts of speech due to morphological richness of Telugu language (\citet{badugu2014morphology}).

After Anusaaraka and several other individual machine translation systems, Indian language to Indian language machine translation (IL-ILMT) project was initiated to work on bidirectional machine translation between nine Indian languages translation. In this consortium-based project, 11 Indian institutes participated and designed a system called Sampark\footnote{\url{http://ilmt.tdil-dc.gov.in/sampark/web/index.php/content}}. This system was developed by building the required `horizontal' and `vertical' resources, tools and other components. This is a transfer-based hybrid machine translation system that includes knowledge-based and statistical modules. So far, it has publicly made available\footnote{\url{https://sampark.iiit.ac.in}} Hindi$\leftrightarrow$Urdu, Hindi$\leftrightarrow$Punjabi, Hindi$\leftrightarrow$Telugu, Hindi$\leftrightarrow$Tamil, Hindi$\leftrightarrow$Bengali, Hindi$\leftrightarrow$Marathi, Hindi$\leftrightarrow$Kannada, Tamil$\leftrightarrow$Telugu, and Tamil$\leftrightarrow$Malayalam machine translation systems. Our continuing work on building machine translation systems for languages of the Purvanchal languages can be considered an extension of this project.

Manipuri (Meitei) was added a few years ago to the list of scheduled languages and belongs to the Sino-Tibetan language family. A recent attempt (\citet{singh2016automatic}) was made to create corpora from two local Manipuri newspapers, written in the Bangla script, and it was made compatible with Unicode through manual correction from Bangla OCR. Similar to Manipuri speech corpora (\citet{jha2010tdil}), the collection of data for Automatic Speech Recognition (ASR) from native speaker(s) and the telephonic speech data transcription was carried out. 

Over the years, NLP and CL community has continued to work on the development of resources for low resource Indian languages (although work on a few languages like Hindi has been going on for more than two decades) and use them for several language-based applications (\citet{jha2010tdil,srivastavainterspeech,kumar2005development}). With the growing use of the Internet in the age of digitization, speech technology (\citet{ mohan2014acoustic}) also came into the picture for Indian languages, and played a crucial role in the development of applications in various domains such as agriculture, health-care, and other services for ordinary people (\citet{kumar2005development}).
Researchers outside India have also contributed to building corpora for Indian languages and one such example is the HindEnCorp\footnote{\url{https://ufal.mff.cuni.cz/hindencorp}}, a monolingual corpus for the Hindi language and a parallel corpus of Hindi-English comprising 787M and 3.8M tokens, respectively (\citet{bojar2014hindencorp}).

For the three Purvanchal languages, the first attempt to collect monolingual Bhojpuri corpus was made through a web crawler that includes dialects, different genres, and domains of Bhojpuri (\citet{singhweb}). Further, this corpus is annotated for POS tagging, and a statistical tagger is developed, which allows us to compare with two other Indo-Aryan languages: Hindi and Oriya (\citet{singh2015statistical, ojha2015training}). A similar attempt (\citet{kumar2011challenges}) was made to collect a monolingual corpus of Magahi language from blogs, stories, magazines, and transcripts of spontaneous conversations carried out in public, private, and home domains, which made it possible to prepare a corpus of around 1 million tokens. POS annotation used BIS standard tagset\footnote{\url{http://tdil-dc.in/tdildcMain/articles/134692Draft\%20POS\%20Tag\%20standard.pdf}}. The tagset consists of 11 coarse-grained tags and around 30 fine-grained tags for each language that it covers. This standard tagset was slightly modified and tested on around 50000 tokens (with fine-grained tags) with TnT, Memory-based tagger, Maximum-entropy tagger, HMM-based tagger, SVM-based tagger by \citet{kumar2012developing}. Maithili corpus from web resources and Wikipedia dump was used towards developing a POS tagger, using CRF, on 52,190 words with an accuracy of 82.67\% (\citet{priyadarshi2020towards}). Apart from this, HMM model was also employed on 10K words with an accuracy of 95.67\% (\citet{Rishikesh2018towards}).

\section{About Bhojpuri, Maithili and Magahi}
\label{bmm-differences}

Bhojpuri belongs to the Indo-Aryan (Indic) language family. Its origin is often attributed to a place in Bihar called Ara/Bhojpur. According to the census of India (2011) data, approximately 50,579,447 speakers use it as their first language. Ironically, a language with such a massive user base is yet to have official recognition and attain the status of a scheduled (or officially recognised) language of India, because the government agencies classify Bhojpuri as a dialect /regional variety of Hindi, in particular of `Eastern Hindi'. The resultant lack of recognition for this language, coupled with low employment opportunities owing to scarce resources for teaching and learning technology has brought tremendous disadvantage to Bhojpuri and its speech community. Setting aside the classifications based partially on socio-political considerations, it is justifiable to treat Bhojpuri as a living language that has a significant user base, and that is spoken in several countries besides India, including Fiji, Nepal, Surinam, and Mauritius. Moreover, its speakers would outnumber speakers of several officially recognized languages and dialects of India, and also of several European languages. Across India, they are spread with high concentrations in the states of Uttar Pradesh, Bihar, Jharkhand, and Chattisgarh.

Magahi became a distinct language around the tenth century, similar to other New Indo-Aryan (NIA) languages. Grierson classified Magahi under the Eastern group of the outer sub-branch. Scholars like Turner have clubbed the Bihari languages with Eastern and Western Hindi (\citet{masica1993indo}). \citet{chatterji1926evolution} gave an entirely different classification, where Western Hindi is almost an isolated group, while Eastern Hindi, Bihari, and other languages are clubbed together in the Eastern group. Nevertheless, the proposition, given above, by \citet{grierson1903linguistic} is the most widely accepted. \citet{jeffers1976syntactic} and others have proposed a similar categorization.

Although Magahi is frequently mislabelled as a dialect of Hindi, the fact is that it is not even part of the same sub-family as Hindi. It was developed from the Maagadhi/Eastern Apbhransha branch of Maagadhi Prakrit. Other languages developing from this branch are Maithili, Bhojpuri, Bangla, Assamese, and Oriya. This deduction is validated using the Grierson's classification and is strengthened even via grammars of Hindi and Magahi. According to the features of the languages, Bangla, Oriya, Assamese, Maithili, Bhojpuri, and Magahi are close cognate sister languages, which means that Bhojpuri and Magahi are sister languages. Consequently, Magahi shares many characteristics with the Bhojpuri language, such as a massive user base of 20.7 million speakers. Maithili is their first cousin, and languages like Bangla, Assamese, and Oriya are their second cousins (\citet{verma1991exploring}). Contemporarily, Magahi is spoken chiefly in the states of Bihar, Jharkhand, West Bengal, and Orissa. 

Maithili serves as the first language to approximately 30 million speakers, who live in the Indian state of Bihar and South-eastern region of Nepal. It is a highly dialect-oriented language having several variations due to social factors such as gender, education level, caste etc. It shows a relatively stronger influence of Sanskrit and, further, has at least ten varieties in India and at least three regional dialects in Nepal. The social status shows the honorific agreement in this language (\citet{yadava2019syntax}).

{\bf Linguistic Features}

Bhojpuri, Maithili and Magahi languages have Subject-Object-Verb (SOV) word order. These languages are written in Devanagari and Kaithi scripts. Like many other modern Indo-Aryan languages these languages are also non-tonal languages. Word-formation in these languages is somewhere between synthetic and analytical typology. These languages, unlike Hindi, lack ergativity. More details about linguistic features of these languages are discussed in the section on annotation issues.

\begin{adjustbox}{valign=t}
\label{typology}
\begin{forest}
[Saurseni Prakrit
    [Saurseni Apbhransha 
        [Braj Bhasha]
        [Kannauji]
        [Khari Boli
            [Western Hindi ]
        ]
    ]
]
\end{forest}
\end{adjustbox}\qquad
\begin{adjustbox}{valign=t}
\begin{forest}
[Magadhan Prakrit
    [Ardha-Magadhi Apbhransha
        [Eastern Hindi]]
    [Magadhi Apbhransha
        [Bihari
            [Bhojpuri]
            [Magahi]
            [Maithili]]
        [Bengali]
        [Oriya]
        [Assamese] ]
]
\end{forest}
\end{adjustbox}

Based on the linguistic features, Bengali, Oriya, Assamese, Maithili, Bhojpuri and Magahi are close cognate languages, which means that Bhojpuri and Magahi are sister languages and Maithili is their first cousin. Languages like Bengali, Assamese and Oriya are their second cousins. Hindi is a distant relative, even though prolonged close contact (with Hindi being the official language of the India and also of the two states where these languages are spoken the most) has made things a bit cloudy. Some examples of the differences are as follows:

\begin{itemize}
    \item \textbf{Non-ergative Construction}: Standard Hindi is an Ergative language. In Hindi when the subject is marked with [ne] (oblique case) then the transitive verb agrees with the object in terms of person, number and gender,  whereas Bhojpuri, Magahi and Maithili are non-ergative languages. For example: \\
\textit{Bhojpuri}: mohan kiwAb paDZalas. \\
\textit{Hindi}: mohan ne kiwAb paDZI. \\
\textit{English Translation}: Mohan read the book. \\
    
    \item \textbf{Emphatic Expressions}: In these languages emphatic particles (like certain case markers) are generally merged with nominals and pronominals, whereas in Hindi we have separate emphatic particles such as [BI], [hI]. For example: \\
\textit{Bhojpuri}: laikav[o] Ayala rahe. \\ 
\textit{Hindi}: ladZakA [BI] AyA WA.  \\
\textit{English Translation}: The boy [too] had come. \\
In this example sentence of Bhojpuri, the emphatic particle [o] is fused with the nominal to show emphasis. \\
\\

    \item \textbf{Classifiers}: Like Bengali and Oriya, Bhojpuri, Magahi and Maithili languages also very often use classifiers with numerals. Classifier markers in these languages are: To, Te, go, Ke, goTa etc., whereas Hindi and its other dialects do not show this characteristic.  For example: \\
\textit{Maithili}: eka [goTa] sajjana. \\
\textit{Hindi}: eka [human-classifier] sajjana. \\
\textit{English Translation}: One [human-classifier] gentlemen. \\

    \item \textbf{Determiners}: Wide use of determiners is again a unique feature of Bhojpuri, Maithili and Magahi languages. In these languages determiners can occur with almost all the proper and common nouns and similar to the emphatic markers, determiners also get fused with nominal forms. This constructional feature marks another point of contrast from Hindi. \\

In Bhojpuri, Maithili and Magahi languages a word ending with [-a] or [-A] sound will take [-vA] suffix, [-i/-I] will take [-yA] suffix and [-u] will take an [-A] suffix. 
\\
Examples: \\
\textit{Bhojpuri}: [XiyavA mAI se kahalas] \\
\textit{Hindi}: [beTI ne mAz se kahA] \\
\textit{English Translation}: Daughter told the mother. \\

\end{itemize}

Bhojpuri, Maithili and Magahi are partially synthetic languages, as can be seen from the above examples. Morphology of these languages differs significantly from the morphology of Hindi in some aspects. One way to perform a preliminary check about the idiosyncrasies of a language is by calculating or applying some statistical measures on the corpus of a language. Before describing this part, we describe text corpus creation for the three languages in the next section.

\section{Text Corpus Creation of Purvanchal Language: Bhojpuri, Maithili and Magahi}

For the project on which this study is based, we collected data from web pages and made a collection of corpus under the supervision of linguists and speakers of these languages. These were qualified linguists and native speakers of these languages, because of which they were also familiar with the writing systems used for these languages. This last part about writing is only partially true, because none of these languages is mediums of education anywhere in India and very few people actually read or write in these languages, even though they are spoken by a large number of people. It is partially true because all these speakers of these three languages are also native speakers of Hindi, which uses the Devanagari script. These speakers often identify themselves as Hindi speakers because Hindi is their main first language, not only for communication, but also for education and official purposes, apart from English, which is still the de-facto first official language of India (although constitutionally, English is the second official language, after Hindi)\footnote{https://rajbhasha.nic.in/en/constitutional-provisions}. None of these speakers, with rare exceptions, regularly reads or writes in these languages.

We collected raw text in Bhojpuri, Magahi and Maithili from online newspapers, magazines, and other relevant resources. The collected data was made balanced and representative to some extent by including varied genres such as sports, editorial, politics, science fiction, entertainment, education, and daily news. In magazines, we preferred [pAwI]\footnote{The items within [] brackets indicate WX notation. The WX (\url{https://en.wikipedia.org/wiki/WX_notation}) notation is used throughout this paper, as is the usual practice for Indian languages for computational purposes.} for Bhojpuri and [axrXaga] novel for Maithili. While building the corpora, we avoided data from blogs so that we might be able to maintain the linguistic authenticity of the sources to the Bhojpuri and Maithili languages. Blog posts in these languages often use a high degree of code-mixing. However, significant deficit of data sources for the Magahi language compelled us to obtain its data from various blogs to at least establish an approximate comparison while making an analysis. Most of sources of corpora on the Web that we used are listed in appendix A.

\subsection{Raw Corpus Cleaning and Issues}

After extracting the corpus from the Web and other sources, we had to clean it in the usual way (such as removing of unwanted characters, symbols), but also, in some cases, to convert the encoding into Unicode (UTF-8). This is because for most Indian languages, Unicode is still not universally used. For the three Purvanchal languages, many sources, particularly those from printed material, were written using a proprietary encoding that has been popular for Desktop Publishing in Devanagari. Since the encoding converter is not perfect (as the mapping between this encoding and Unicode is not straightforward), there were some errors in conversion to Unicode. Because of this, some `noise' is present in the text corpus like extra punctuation, invalid ASCII characters, foreign languages words, misspelled words etc. Part of the cleaning process involved removing such noise as far as possible, given the time and resource constraints. We also removed the code mixed sentences from the crawled data, only excluding sentences with Hindi nouns, after verifying with a language-specific linguist. There are still some errors left that were introduced due to encoding conversion.

Since there were not many significant lexical or structural mistakes, so the corpus cleaning process, on the whole, was not very time-consuming, nor could we afford to spend too much time on it. Part of the problem was the lack of standardisation and the confusability among the three language. We have described this in more detail in the section on language identification. There were also time and resource constraints.  We found several minor orthographic issues in the collected corpora of these three languages, such as for [Sa, Ra]. An example of words where such problems were found is: [akCara  $\rightarrow$ akRara] in Maithili. The problem in these cases is that the orthographic practices of Hindi are carried over to the concerned languages, causing mistakes, e.g. [akCara] is written as [akSar], which is the Hindi spelling for this word, roughly  meaning orthographic syllable. Another common problem is due to the diacritic called Avagrah. Avagrah is frequently used in Bhojpuri, Magahi and Maithili, and it indicates lengthening of the preceding vowel for our purposes. In many cases, this diacritic was missing, which can change the meaning of a word in these languages. We tried to introduce it in places where it was missing, but this process may not have been rigorous due to the lack of available annotator-time.

Since these languages (previously considered dialects) do not have much of standardization, the words are not spelled consistently, adding to the problem of language processing. A word can have multiple spelling variations. Some examples of problems due to these reasons are given below:

\begin {itemize}
   \item{\textbf{Orthographic variations}}: As mentioned above, words having the same meaning can often be spelled in more than one ways, i.e., there is a lot of orthographic variation. This is due to the lack of standardization for the three languages. This problem is present, to a much lesser extent, even in Hindi. For example, consider one form of variants formed by replacing Anusvara with Anunasik or Anusvara with half-nasal consonants (such as in [samvawa/saMvawa], [samBala/saMBala/saBala]. Examples of other orthographic variants may include [karie/kariye (e/ye)], [karAyI/karAI, uTAyI/uTAI (I/yI)], [bAyIM/bAyIZ], [karyavAI/kAryavAhI/kAravAhI], [BEyA/BaiA/BaiyA], [vAyasarAya/vAisarAya] and [suwala/sowala].\\
   
    \item {\textbf{Particle identification}}: Particles are functional words which do not have very clear meanings, even as function words. Examples of such particles include [vA, bA] and  [mA] constructions. Bhojpuri and Magahi have two types of roots: one without [vA] and another attached with [vA], e.g. [laikA] (boy), and [laIkavA] occur in masculine and non-masculine words, but in feminine words, [yA] suffix is present. Maithili has a similar type of construction, however, it uses [mA] and [bA] suffixes. For example: [CoramA] (boy) and [CorabA]. It is debatable how these suffixes change the meanings. Strictly speaking, these are suffixes and are not words, so they should not be called particles in the sense Hindi has particles such as [hI] or [wo]. Still, they pose problems while annotating the corpus.\\

    \item {\textbf{Incorrectly spelled words or foreign language words}}: As some words may not be correct or borrowed from any different language than the chosen three languages for analyses, they are often not be spelled correctly (or consistently) as there is no standard spelling. Standard spelling is a general problem for all three Purvanchal languages even for their native words, but it is particularly problematic for borrowed words. For the subset of data that was annotated, we tried to filter out such problematic words to some extent. \\
    
    \item{\textbf{Numbers, Abbreviations, Symbols}}: We also had to consider the use of alternative written forms for words such as [SrI, dA.], symbols (*, \& )  and numbers (3224.34).
\end{itemize}

\subsection{Frequency Distribution Analysis}
\label{sec4_2}

Statistical analyses of the corpora can help us understand the language characteristics. In quantitative analysis, we classify different linguistic properties or features of a particular language, count them, and construct more complex statistical models in an attempt to explain what is observed. It allows us to discover and differentiate the genuine from the improbable reflections of phenomena capable of describing the behaviors of a language or a variety of languages. On the other hand, the quantitative analysis aims at providing a complete and detailed description of the observed phenomena obtained from the corpus.

It may be of additional interest to the reader that the work described by \citet{kumar2007statistical} involved corpora analyses on many other Indian languages, such as Hindi, Telugu, etc., one or all of them could be used for comparison with the three primary languages of interest in this paper, i.e., Bhojpuri, Maithili, and Magahi. In our work, wherever a comparison is useful and possible, we have augmented our analysis in the following sections by comparing the measures and metrics with the Hindi language, similar to the work by \citet{kumar2007statistical}. A monolingual Hindi corpus\footnote{\url{http://www.cfilt.iitb.ac.in/Downloads.html}} (Europarl v7), has been used for this comparison.

\subsubsection{\textbf{Unigram}}
From the corpus, the initial statistical analysis in terms of the most basic units of language can be performed, e.g. word frequency, vocabulary size etc.. We extracted the total number of sentences, tokens, types, and rare types. Here, the type can be defined as the vocabulary or lexical variety of the language expressed as a number of distinct (unique) tokens available in the corpus. Further, the rare types are those tokens which occur only once throughout the corpus, i.e., with frequency 1. The relationship between the types and tokens is sometimes determined by calculating the type-token ratio (TTR) that shows how often new word forms appear in the corpus (\citet{sarkar2004easy}).
Since the TTR is sensitive to the corpus size, it has further been improved to avoid the effect of corpus size by \citet{doi:10.1080/09296171003643098}. This better measure, namely moving-average type-token ratio (MATTR), which is less dependent on the sentence length or corpus size, since it works on a sliding window for a fixed length. We have performed the MATTR experiments with 500 window size (throughout this paper), as suggested by \citet{doi:10.1080/09296171003643098}.

Intuitively, MATTR also roughly indicates the vocabulary size, and therefore, the morphological complexity or richness. This is partly because we are counting over word forms and not lexemes or `root words'. We can examine the relationship between the lexical or morphological richness and the TTR, and the MATTR values: high TTR, and MATTR meaning high lexical/morphological richness and vice-versa. With reference to these terms in the context of the corpus that we created of three Purvanchal languages, the Table~\ref{unigram analysis} lists the total the number of sentences, tokens, types, rare types, and also the TTR, and MATTR corresponding to each language.

\begin{table}[!ht]
\caption{Unigram analysis of Purvanchal Languages, compared with Hindi. Here * indicates the TTR ratio over the number of sentences according to the Maithili corpus, which was the smallest. TTR is calculated on this (minimum) corpus size for all the four languages, to observe the effect of corpus size on TTR.}
\label{unigram analysis}
\begin{tabular}{cccccccl}
\toprule
\textbf{Language}	&\textbf{Total Sentences}	& \textbf{Total Tokens}	& \textbf{Total Types} & \textbf{Rare Types} & \textbf{TTR} & \textbf{MATTR} &  \textbf{TTR*}\\
\midrule
Bhojpuri &	51374 &	704543 &	38680 &	178 &	5.49 & 0.572 & 9.71\\
Maithili &	19948 &	267881 &	27043 &	10597 &	10.09 & 0.587 & 10.09\\
Magahi &	31251 &	424434 &	35253 &	12978 &	8.30 & 0.575 & 9.09\\
Hindi & 221768 & 2917271 & 151516 & 89436 & 5.19 & 0.516 & 11.40\\
\bottomrule
\end{tabular}
\end{table}
 
From the Table~\ref{unigram analysis}, we find that the value of TTR, and MATTR comes out to be highest for the Maithili language among the three languages. This indicates that Maithili and Hindi are the most and the least lexically/morphologically rich, respectively, among the four languages. However, a look at the second column tells us that the number of sentences for the Maithili language is the least and quite less in comparison to the remaining three languages. Thus, only from Table~\ref{unigram analysis}, we cannot rule out the possibility that the value of TTR (if not MATTR) could have been higher had we created a larger corpus of the Maithili language, comparable to the other three languages. 

While this is plausible, the previously available linguistic knowledge about the complexities or richness of the four languages at various linguistic levels does not agree with this observation, based on the last column. This knowledge agrees more with the TTR (without the same corpus size restriction), and the MATTR column . This could be interpreted in two ways. Either that the corpus size should not be the same for all languages for the TTR to be compared for lexical/morphological richness, or that the quantitative approach has some limitations, because intuitively the same corpus size restriction seems reasonable for quantitative analysis. However, MATTR values agree with the linguistic knowledge.

\begin{table}[!ht]
\caption{Percentage of Types covering $n$ percent of the corpus where $n$ can be 50, 60, 70, 80, and 90. Here * indicates the type coverage over the percent of the corpus according to the Maithili corpus, which was the smallest.}
\label{ttr ratio}
\begin{tabular}{ccccccccccccc}
\toprule
\textbf{Language}	&\textbf{50\%} &\textbf{50\%*}	& \textbf{60\% } &\textbf{60\%*}	& \textbf{70\%} &\textbf{70\%*}	& \textbf{80\%} &\textbf{80\%*}	 & \textbf{90\%} &\textbf{90\%*} \\
\midrule
Bhojpuri & 60.15 & 89.71 & 70.49 & 99.88 & 79.36 & 99.89 & 85.25 & 99.97 & 88.47 & 99.99 \\
Maithili & 66.79 & 66.79 & 76.82 & 76.82 & 85.13 & 85.13 & 89.49 & 89.49 & 93.61 & 93.61 \\
Magahi & 71.49 & 56.79 & 74.76 & 62.83 & 80.6 & 70.81 & 86.64 & 81.92 & 93.09 & 91.69 \\
Hindi & 52.4 & 60.27 & 61.67 & 68.46 & 69.15 & 76.06 & 77.44 & 84.41 & 92.49 & 90.56 \\
\bottomrule
\end{tabular}
\end{table}

In order to further explore this inference, we come to our second step of statistical analysis, provided in Table~\ref{ttr ratio}, that is, the percentage of types required to cover a certain percentage ($n$) of the random corpus, where $n$ can be 50, 60, 70, 80, and 90 (100\% corpus coverage is already done in Table~\ref{unigram analysis}). For instance, the value in the third row and fourth column shows a value of 74.76, which means that out of total 35253 types available in the entire corpus for the Magahi language, only 74.76\% types (or 26355 types approximately) are available in the 60\% tokens (or 254660 tokens nearly) chosen randomly out of the entire number of tokens (424434) in the corpus. We have closely followed settings of ~\citet{bharati2000basic}. For all values of $n$, it can be observed that the percentage of corpus coverage is highest for Bhojpuri. Similarly, except for the case when Magahi exceeds Maithili in percentage coverage of types, Maithili consistently exhibits the highest percentage coverage of the types. These results seem to contradict the conclusion from the second last column of Table~\ref{ttr ratio} (TTR on full corpus size) as well as previous linguistic knowledge. However, these results in Table~\ref{ttr ratio} are on the same percentages, not on the same corpus sizes in terms of the number of tokens. Since different languages have different degrees of lexical richness, they will need different amounts of corpora to have the same coverage of types. Therefore, it seems better to compare the MATTR in the second last column of Table~\ref{unigram analysis}, rather than the last column.

\begin{table}
\caption{The top ten most frequent words for the three languages in decreasing order of their frequency (Freq) with their Relative frequency (RF) and Cumulative Coverage (CC) values, compared with Hindi}
\label{frequent word}
    \begin{tabular}{ccc}
        \centering
        \begin{tabular}{c|ccc}
        \toprule
        \textbf{Bhojpuri}	&	\textbf{Freq}	&	\textbf{RF}	& \textbf{CC} \\    \hline
        [ke] & 49700 & 0.070 & 0.070 \\ \newline
        [meM] & 14893 & 0.021 & 0.091 \\ \newline
        [A] & 12990 & 0.018 & 0.109 \\ \newline
        [se] & 12893 & 0.018 & 0.127 \\ \newline
        [bA] & 9386 & 0.013 & 0.140 \\ \newline
        [kA] & 8279 & 0.011 & 0.151 \\ \newline
        [nA] & 7304 & 0.010 & 0.161 \\ \newline
        [wa] & 7256 & 0.010 & 0.171 \\ \newline
        [ki] & 5849 & 0.008 & 0.179 \\ \newline
        [rahe] & 5839 & 0.008 & 0.187  \\
        \bottomrule
       \end{tabular} & &
      
        \begin{tabular}{c|ccc}
        \toprule
        \textbf{Maithili}	&	\textbf{Freq}	&	\textbf{RF}	& \textbf{CC} \\
        \hline
        [aCi]  &	7359 &	0.027 &	0.027 \\ \newline
        [saBa]  &	4021 &	0.015 &	0.042 \\ \newline
        [je]  &	3797 &	0.014 &	0.056 \\ \newline
        [nahi]  &	3318 &	0.012 &	0.068 \\ \newline
        [o]  &	2685 &	0.01 &	0.078 \\ \newline
        [ka]  &	2281 &	0.008 &	0.086 \\ \newline
        [me]  &	2270 &	0.008 &	0.094 \\ \newline
        [lela]  &	2128 &	0.007 &	0.101 \\ \newline
        [gela]  &	2014 &	0.007 &	0.108 \\ \newline
        [I]  &	1906 &	0.007 &	0.115 \\
        \bottomrule
       \end{tabular} \\ \\ \\
       
        \begin{tabular}{c|ccc}
        \toprule
        \textbf{Magahi}	&	\textbf{Freq}	&	\textbf{RF}	& \textbf{CC} \\
        \hline
        [ke] 	 & 	29977	 & 	0.07	 & 	0.07 \\ \newline
        [meM] 	 & 	10050	 & 	0.023	 & 	0.093 \\ \newline
        [he] 	 & 	7603	 & 	0.017	 & 	0.11 \\ \newline
        [se] 	 & 	7521	 & 	0.017	 & 	0.127 \\ \newline
        [hala] 	 & 	6465	 & 	0.015	 & 	0.142 \\ \newline
        [Au] 	 & 	5651	 & 	0.013	 & 	0.155 \\ \newline
        [para] 	 & 	3998	 & 	0.009	 & 	0.164 \\ \newline
        [gela] 	 & 	3598	 & 	0.008	 & 	0.172 \\ \newline
        [ki] 	 & 	2591	 & 	0.006	 & 	0.178 \\ \newline
        [I] 	 & 	2585	 & 	0.006	 & 	0.184 \\
        \bottomrule
       \end{tabular} & &
       
        \begin{tabular}{c|ccc}
        \toprule
        \textbf{Hindi}	&	\textbf{Freq}	&	\textbf{RF}	& \textbf{CC} \\
        \hline
        [ke] 	 & 	102401	 & 	0.0351	 & 	0.0351 \\ \newline
        [hE] 	 & 	87065	 & 	0.0298	 & 	0.0649 \\ \newline
        [meM] 	 & 	80276	 & 	0.0275	 & 	0.0924 \\ \newline
        [kI] 	 & 	66900	 & 	0.0229	 & 	0.1153 \\ \newline
        [se] 	 & 	48278	 & 	0.0165	 & 	0.1318 \\ \newline
        [Ora] 	 & 	47449	 & 	0.0163	 & 	0.1481 \\ \newline
        [kA] 	 & 	47017	 & 	0.0161	 & 	0.1642 \\ \newline
        [ko] 	 & 	40774	 & 	0.014	 & 	0.1782 \\ \newline
        [para] 	 & 	26402	 & 	0.0091	 & 	0.1977 \\ \newline
        [BI] 	 & 	25405	 & 	0.0087	 & 	0.2064 \\
        \bottomrule
       \end{tabular} \\
    \end{tabular}
\end{table}

Taking this analysis further, we summarize in Table ~\ref{frequent word} the Cumulative Coverage, the Frequency and Relative Frequency (or probability) of the occurrence of tokens for a particular language. This table shows the top ten most frequent tokens in the three languages in decreasing order of their frequencies (or RF). A high value of CC indicates that the probability of occurrence regarding both rank and frequency is high.

We first observe the relationship between Bhojpuri and Magahi using various patterns of data in Table ~\ref{frequent word}. The token [ke] exhibits the highest frequency in both these languages. Further, the CC of the word [ke] is equal in Magahi in comparison to Bhojpuri. Besides, this word does not exist in this list of the Maithili language, indicating that there is a closer association between Magahi and Bhojpuri. Similar to [ke], [meM], [se] and [ki] are on the equally high frequent ranks in Bhojpuri and Magahi. The CCs are again found to be higher in Magahi (0.093, 0.127, 0.178) than Bhojpuri (0.091, 0.127, 0.179). Thus, since phylogenetically related languages have  a certain degree of similarity in terms of frequent words by order and rank, the four tokens considered above strengthen an assertion that Bhojpuri and Magahi are genetically related languages. 

Additionally, Maithili and Magahi also have interesting patterns in terms of frequency and CC. Consider two of the most frequent words in the list: [I] and [gela]. Firstly, these words do not appear in Bhojpuri's top ten most frequent words list, but occur at the same rank of occurrence among the two languages. Further, the [gela] word is nearly equally frequent in Magahi and Maithili, and differs by just one rank between the two languages. The CCs of both of the words in Magahi is higher, being 1.184 and 0.172, while it is 0.115 and 0.108 in Maithili, respectively. Similar to the assertion in the preceding paragraph, our assertion of genetic relations between Maithili and Magahi languages is strengthened. However, it also indicates that perhaps Bhojpuri and Magahi are closer than Magahi and Maithili.

Finally, looking at the total CC for the top ten words, the value is significantly lower for Maithili than it is for Bhojpuri and Magahi, even though Maithili has the least amount of corpus. This value is the highest for Hindi, but that is most probably due to the large size of Hindi corpus and also the fact that Hindi is used much more for academic purposes, thereby having a larger vocabulary. It also supports the linguistic intuition that Maithili is lexically or morphologically richer.

\subsubsection{\textbf{Character}}
\begin{figure}[!ht]
\centering
\includegraphics[width=5in]{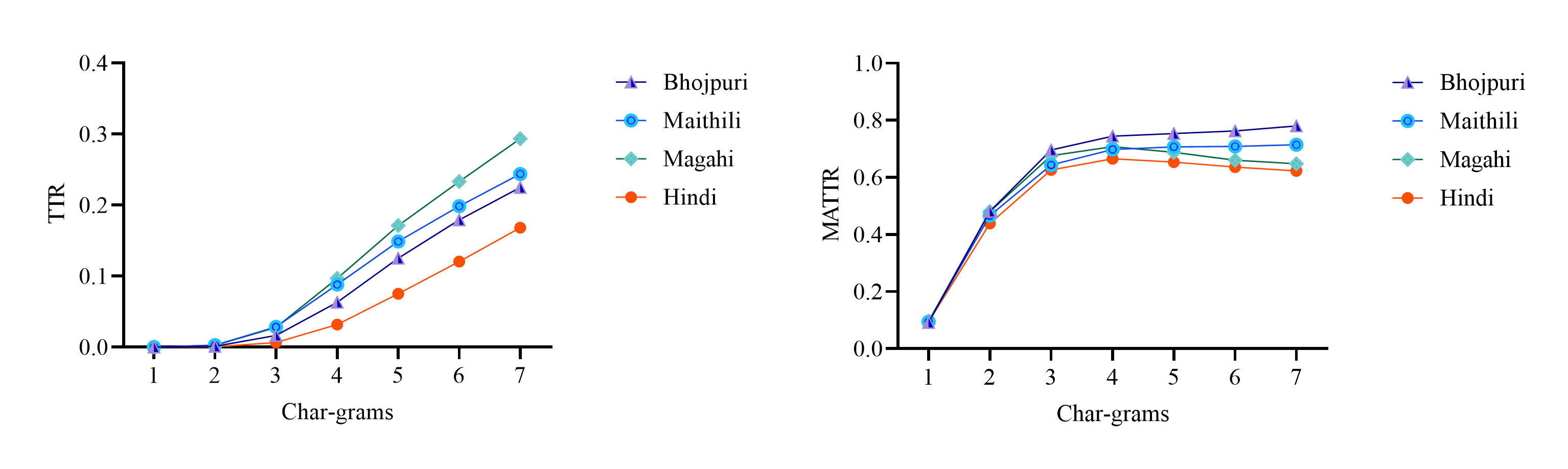}
\caption{TTR and MATTR at character level of 1-7 grams for each language}
\label{character level analysis}
\end{figure}

Since three languages, and also Hindi, use the Devanagari script almost in the same way (i.e., writing systems are also similar), comparing statistics about characters used in the corpora can also perhaps give some insights, or at least validate (or invalidate) some linguistic intuition about these languages, we have performed an exercise similar to words at the character level. This is also supported by the fact that the Devanagari script has an almost one-to-one mapping from letters to phonemes. We have considered character $n$-grams (excluding space and special symbols), with $n$ up to 7, as seven n-grams of characters seem to capture the identity of a language quite well~(\citet{singh2006study}) for language identification. Figure \ref{character level analysis} lists the TTR and MATTR values for $n$-grams for 1 to 7 levels for the four languages. For character unigrams, the values are nearly zeros for all the languages, because the number of unigrams is quite small (equal to the size of the Devanagari character set), while the total number of character tokens is much larger. This begins to change from bigram upwards. The trends, in terms of numbers, seem to run opposite to those for word level TTRs. We interpret it as meaning that higher values of character $n$-gram TTR and MATTR indicate greater lexical richness (relative vocabulary size), but lesser morphological complexity, which is in consonance with linguistic intuition. In phonological terms, it could also mean that higher TTR and MATTR for character (or phoneme) $n$-grams is indicative of greater phonological variety. We are not aware of any linguistic intuition about this, but perhaps phonologists familiar with these languages can comment on this better.

\begin{table}[!ht]
\caption{Frequency of orthographic syllables based on positions}
\label{orthographic1 syllable}
\centering
\begin{tabular}{cccccccc}
\toprule
\textbf{Language}	&	\textbf{Total syllables}& \textbf{Initial}	 & \textbf{Percent}	&	\textbf{Medial} &	\textbf{Percent} & \textbf{Final} & \textbf{Percent} \\
\midrule
Bhojpuri & 85588 & 25870 & 30.22\% & 34536 & 40.36\% & 25182 & 29.42\%\\
Maithili & 99218 & 27001 & 27.21\%  & 45608 & 45.96\% & 26609 & 26.82\%\\
Magahi & 74712 & 23120 & 30.95\% &  28824 & 38.58\% & 22768 & 30.47\%\\
Hindi & 113318 & 32490 & 28.67\% &  49817 & 43.96\% &  31011 & 27.37\%\\
\bottomrule
\end{tabular}
\end{table}

\begin{table}[!ht]
    \caption{TTR and MATTR over the unigram syllable, Here * indicates the TTR over the equal number of sentences according to Maithili corpus}
    \label{syllable_ttr}
    \centering
    \begin{tabular}{lllll}
    \toprule
    & \textbf{Bhojpuri} & \textbf{Maithili} & \textbf{Magahi} & \textbf{Hindi}\\
    \midrule
    TTR & 0.1466 & 0.2945 & 0.1964 & 0.1795 \\
    TTR* & 0.3338 & 0.2945 & 0.2711 & 0.6052\\
    MATTR & 0.270 & 0.271 & 0.265 & 0.303 \\
    \bottomrule
    \end{tabular}
\end{table}

\subsubsection{\textbf{Syllable or Akshara}}
Since there is an almost direct mapping from letters to phonemes in Devanagari, it is also possible to analyze the data in terms of orthographic syllables (which are very similar to phonological syllables). The term Akshara is used for such orthographic syllables. This term has been used since the time of the Paninian grammarians, but in our case, we use a slightly modified version of syllable that does not necessarily end with a vowel (like traditional Akshara does). This modified term came into use perhaps due to the requirements of building computing technology for Indic languages, which use Brahmi-derived scripts\footnote{\url{http://www.tdil-dc.in/tdildcMain/articles/737061Hindi\%20Script\%20Grammar\%20ver\%201.4-2.pdf}}.

In this section, we perform a similar exercise as in the preceding two sections, but based on the level of orthographic syllables or Aksharas. It has been observed and reported by multiple authors that Akshara is a better linguistic unit than character for many language processing and analysis purposes for Indian languages~(\citet{singh2006computational,kumar2007statistical}). The grammar for orthographic syllabification is based on the document defining the ISCII standard\footnote{\url{http://varamozhi.sourceforge.net/iscii91.pdf}}.

In general, the syllable structure is defined by the onset-nucleus-coda (consonant-vowel-consonant). The core component of the syllable is the nucleus that is a vowel, and the remaining components -- onset and coda -- are optional, and are the preceding and the following consonant, respectively. An orthographic syllable for us is an orthographic unit of the Devanagari script (conjuncts included) and which may or may not end with a vowel, but does correspond to one or more phonological syllables. Since the position of a syllable in a word (initial, medial or final) is also sometimes important, we consider the numbers of orthographic syllables for each of these positions in Table~\ref{orthographic1 syllable}. The TTR, and MATTR value for orthographic syllables are given in Table \ref{syllable_ttr}. The top ten most frequent orthographic syllables in Bhojpuri, Maithili, Magahi, and Hindi languages are given in Table \ref{orthographic syllable}.

Observations from Table \ref{orthographic1 syllable} are as follows. The percentages of syllables that occur in the initial and the final are almost the same in all the four languages. The percentages of syllables that can occur in a medial position is much larger for all languages, which is natural since the medial part of the word can contain more than one syllables and most words are more than three syllables long. The differences in the percent of syllables that occur in the medial position for each of these languages do tell us indirectly about the complexity ~(\citet{maddieson2009calculating}). Once again, Maithili has the highest percentage of syllables that occur in the medial position, meaning it has longer and more complex words, in spite of its corpus being the smallest. Hindi also has a large percentage of syllables that occur in the medial, but that is most probably due to the fact that it is used for academic purposes and has a lot of recently borrowed technical or semi-technical long words originating from Sanskrit. Such words are not normally used in spoken languages, whereas the words in the other three languages (as per our corpus) are used mostly in spoken language, since these languages are not used much in the written form, and therefore, for academic purposes. The same also applies to phonological complexities for these languages. In fact, it is difficult to separate phonological complexity from morphological complexity solely based on these statistics.

Table \ref{syllable_ttr} further strengthens the above observation about phonological complexity of the four languages. Earlier TTR and MATTR have been used for morphological complexity while working at the word level ~(\citet{kettunen2014can}). It seems to be a reasonable assumption that if these measures are used at the syllable level, they will similarly reflect phonological complexity since we are dealing with different kinds of symbolic complexities.

Table \ref{orthographic syllable} is given mainly for reference, although it does not seem to be very informative, except in showing that Hindi is lexically richer than the other languages, which can be explained by the reasons mentioned above.

\begin{table}
\caption{Top-10 most frequent orthographic syllable}
\label{orthographic syllable}
    \begin{tabular}{ccc}
        \centering
        \begin{tabular}{c|ccc}
        \toprule
        \textbf{Bhojpuri}	&	\textbf{Freq}	&	\textbf{RF}	& \textbf{CC} \\    \hline
        [ra]  	 & 	30411	 & 	0.052	 & 	0.052 \\ \newline
        [ke] 	 & 	21698	 & 	0.037	 & 	0.089 \\ \newline
        [ka]  	 & 	21491	 & 	0.037	 & 	0.126 \\ \newline
        [la]  	 & 	21144	 & 	0.036	 & 	0.162 \\ \newline
        [na]  	 & 	20058	 & 	0.034	 & 	0.196 \\ \newline
        [wa]  	 & 	17504	 & 	0.03	 & 	0.226 \\ \newline
        [ha]  	 & 	16540	 & 	0.028	 & 	0.254 \\ \newline
        [i]  	 & 	13746	 & 	0.023	 & 	0.277 \\ \newline
        [sa]  	 & 	12295	 & 	0.021	 & 	0.298 \\ \newline
        [pa]  	 & 	11912	 & 	0.02	 & 	0.318 \\
        \bottomrule
       \end{tabular} & &
      
        \begin{tabular}{c|ccc}
        \toprule
        \textbf{Maithili}	&	\textbf{Freq}	&	\textbf{RF}	& \textbf{CC} \\
        \hline
        [ka] 	 & 	48100	 & 	0.066	 & 	0.066 \\ \newline
        [ra]  	 & 	34284	 & 	0.047	 & 	0.113 \\ \newline
        [la]  	 & 	27753	 & 	0.038	 & 	0.151 \\ \newline
        [na]  	 & 	23358	 & 	0.032	 & 	0.183 \\ \newline
        [wa]  	 & 	20252	 & 	0.028	 & 	0.211 \\ \newline
        [sa] 	 & 	18977	 & 	0.026	 & 	0.237 \\ \newline
        [a]  	 & 	17502	 & 	0.024	 & 	0.261 \\ \newline
        [ha]  	 & 	16548	 & 	0.023	 & 	0.284 \\ \newline
        [pa] 	 & 	14810	 & 	0.02	 & 	0.304 \\ \newline
        [ya]  	 & 	14495	 & 	0.02	 & 	0.324 \\
        \bottomrule
       \end{tabular} \\ \\ \\
       
        \begin{tabular}{c|ccc}
        \toprule
        \textbf{Magahi}	&	\textbf{Freq}	&	\textbf{RF}	& \textbf{CC} \\
        \hline
        [la]  	 & 	34101	 & 	0.058	 & 	0.058 \\ \newline
        [ra]  	 & 	33885	 & 	0.057	 & 	0.115 \\ \newline
        [na]  	 & 	25109	 & 	0.042	 & 	0.157 \\ \newline
        [ha]  	 & 	20885	 & 	0.035	 & 	0.192 \\ \newline
        [ka]  	 & 	20811	 & 	0.035	 & 	0.227 \\ \newline
        [ke]  	 & 	20317	 & 	0.034	 & 	0.261 \\ \newline
        [wa]  	 & 	12922	 & 	0.022	 & 	0.283 \\ \newline
        [ma]  	 & 	12864	 & 	0.022	 & 	0.305 \\ \newline
        [sa]  	 & 	12155	 & 	0.021	 & 	0.326 \\ \newline
        [pa]  	 & 	11464	 & 	0.019	 & 	0.345 \\
        \bottomrule
       \end{tabular} & &
       
        \begin{tabular}{c|ccc}
        \toprule
        \textbf{Hindi}	&	\textbf{Freq}	&	\textbf{RF}	& \textbf{CC} \\
        \hline
        [ra] 	 & 	336035	 & 	0.0518	 & 	0.0518 \\ \newline
        [ka] 	 & 	228890	 & 	0.0353	 & 	0.0871 \\ \newline
        [na] 	 & 	179069	 & 	0.0276	 & 	0.1147 \\ \newline
        [sa]	 & 	177773	 & 	0.0274	 & 	0.1421 \\ \newline
        [pa] 	 & 	138006	 & 	0.0213	 & 	0.1634 \\ \newline
        [ke] 	 & 	131131	 & 	0.0202	 & 	0.1836 \\ \newline
        [wa] 	 & 	127159	 & 	0.0196	 & 	0.2032 \\ \newline
        [kA] 	 & 	119209	 & 	0.0184	 & 	0.2216 \\ \newline
        [a] 	 & 	105388	 & 	0.0163	 & 	0.2379 \\ \newline
        [ha] 	 & 	105306	 & 	0.0162	 & 	0.2541 \\
        \bottomrule
       \end{tabular} \\
    \end{tabular}
\end{table}

\subsubsection{\textbf{Morpheme}}
Since we are interested in relative morphological complexities of the four languages, performing the same exercise as in the previous three sections for morphemes should be more reliably informative. Since we do not have a completely accurate algorithm or grammar for segmenting words into morphemes, we employ the Morfessor tool~(\citet{smit2014morfessor}) for obtaining morphemes from the corpora. This tool uses a machine learning based algorithm that segments the words into morphemes, based on the corpus provided.  Morfessor has two phases: training and decoding. The training data (unannotated word set: $D_w$) is used as the input for the training phase, and the log-likelihood cost estimation function learns the theta-model parameter, defined by $L(D_w, \theta)$. The learned parameter $\theta$ is used for morphological segmentation of new `test' data ($W$) in the decoding phase.

\begin{equation}
    Training:  argmax_\theta L(D_w, \theta)
\end{equation}{}
\begin{equation}
    Decoding: A = \varphi(W;\theta)
\end{equation}{}

\begin{table}
\caption{Top-10 frequent morphemes enlisted in decreasing order of Frequency/Relative Frequency}
\label{morpheme frequency}
    \begin{tabular}{ccc}
        \centering
        \begin{tabular}{c|ccc}
        \toprule
        \textbf{Bhojpuri}	&	\textbf{Freq}	&	\textbf{RF}	& \textbf{CC} \\    \hline
        [ke]	 & 	53802	 & 	0.051	 & 	0.051 \\ \newline
        [wa]	 & 	15628	 & 	0.015	 & 	0.153 \\ \newline
        [meM]	 & 	15096	 & 	0.014	 & 	0.181 \\ \newline
        [A]	 & 	14910	 & 	0.014	 & 	0.195 \\ \newline
        [se]	 & 	13871	 & 	0.013	 & 	0.208 \\ \newline
        [na]	 & 	12627	 & 	0.012	 & 	0.233 \\ \newline
        [kA]	 & 	10820	 & 	0.01	 & 	0.254 \\ \newline
        [bA]	 & 	10377	 & 	0.01	 & 	0.264 \\ \newline
        [nA]	 & 	9903	 & 	0.009	 & 	0.273 \\ \newline
        [la]	 & 	8169	 & 	0.008	 & 	0.29 \\
        \bottomrule
       \end{tabular} & &
      
        \begin{tabular}{c|ccc}
        \toprule
        \textbf{Maithili}	&	\textbf{Freq}	&	\textbf{RF}	& \textbf{CC} \\
        \hline
        [ka]	 & 	23537	 & 	0.055	 & 	0.055 \\ \newline
        [me]	 & 	8788	 & 	0.021	 & 	0.144 \\ \newline
        [aCi]	 & 	7383	 & 	0.017	 & 	0.161 \\ \newline
        [ke]	 & 	6581	 & 	0.015	 & 	0.176 \\ \newline
        [saz]	 & 	5733	 & 	0.013	 & 	0.189 \\ \newline
        [saBa]	 & 	5659	 & 	0.013	 & 	0.202 \\ \newline
        [je]	 & 	3905	 & 	0.009	 & 	0.232 \\ \newline
        [nahi]	 & 	3353	 & 	0.008	 & 	0.24 \\ \newline
        [o]	 & 	3130	 & 	0.007	 & 	0.247 \\ \newline
        [para]	 & 	2725	 & 	0.006	 & 	0.253 \\
        \bottomrule
       \end{tabular} \\ \\ \\
       
        \begin{tabular}{c|ccc}
        \toprule
        \textbf{Magahi}	&	\textbf{Freq}	&	\textbf{RF}	& \textbf{CC} \\
        \hline
        [ke]	 & 	33935	 & 	0.048	 & 	0.048 \\ \newline
        [meM]	 & 	10145	 & 	0.014	 & 	0.184 \\ \newline
        [la]	 & 	9061	 & 	0.013	 & 	0.223 \\ \newline
        [he]	 & 	8694	 & 	0.012	 & 	0.235 \\ \newline
        [se]	 & 	8250	 & 	0.012	 & 	0.247 \\ \newline
        [na]	 & 	7639	 & 	0.011	 & 	0.28 \\ \newline
        [hala]	 & 	6952	 & 	0.01	 & 	0.29 \\ \newline
        [Au]	 & 	5663	 & 	0.008	 & 	0.298 \\ \newline
        [ho]	 & 	5241	 & 	0.007	 & 	0.305 \\ \newline
        [gela]	 & 	5140	 & 	0.007	 & 	0.312 \\
        \bottomrule
       \end{tabular} & &
       
        \begin{tabular}{c|ccc}
        \toprule
        \textbf{Hindi}	&	\textbf{Freq}	&	\textbf{RF}	& \textbf{CC} \\
        \hline
        [ke]	 & 	24053	 & 	0.028	 & 	0.028 \\ \newline
        [meM]	 & 	20465	 & 	0.024	 & 	0.052 \\ \newline
        [hE]	 & 	17378	 & 	0.02	 & 	0.072 \\ \newline
        [kI]	 & 	16957	 & 	0.02	 & 	0.092 \\ \newline
        [se]	 & 	13457	 & 	0.016	 & 	0.108 \\ \newline
        [Ora]	 & 	12408	 & 	0.014	 & 	0.122 \\ \newline
        [kA]	 & 	12287	 & 	0.014	 & 	0.136 \\ \newline
        [ko]	 & 	10476	 & 	0.012	 & 	0.148 \\ \newline
        [ne]	 & 	9457	 & 	0.011	 & 	0.159 \\ \newline
        [para]	 & 	7393	 & 	0.009	 & 	0.168 \\
        \bottomrule
       \end{tabular} \\
    \end{tabular}
\end{table}

We used the Morfessor 
2.0\footnote{\url{https://morfessor.readthedocs.io/en/latest/}} version, with unsupervised learning. Each word in the data is re-written as a sequence of its constituent morphs (hopefully approximating morphemes). This section provides statistics similar to the preceding sections, but at the morpheme level, as given in the Table \ref{morpheme frequency}.

Apart from its purpose as a reference, this table provides two main takeaways. Frequencies and relative frequencies (RFs) do not, by themselves, provide much information that is generalisable, although it may be of interest to linguists who want to study these languages. For our purposes, the first takeaway is very similar to the one from preceding tables. In other words, from the cumulative coverage (CC) values, we can observe the same trend that Maithili has the higher morphological complexity among the three Purvanchal languages. Although Hindi again has a high value, it can be explained in the same way as earlier, i.e., larger and more complex technical or academic vocabulary due to its usage for academic purposes and as a standard language at the national level.

An observation is that the top ten morphemes are very less similar (phonologically and orthographically) for Maithili, as compared to the other languages, which also matches the existing linguistic knowledge about the typological nature of Maithili. The popular knowledge about Maithili says that it is, in fact, closer to Bangla than to Hindi (\citet{chatterji1986origin}). This suggests that it might be a good idea for future work to compare these three languages with Bangla (for more clarity, please see Section \ref{bmm-differences}).

In fact, there are several common morphemes ([meM], [ke], [se], [na], [nA]) among the top ten for Bhojpuri, Magahi and Hindi, but none of them occurs in the list for Maithili. This might be the strongest evidence of the common knowledge that Maithili is typologically the most distinct among these four languages.

\subsubsection{\textbf{Zipf's curve}}
Since all languages  have a long tail distribution in terms of frequencies and ranks of the words in the vocabulary, in this section we describe the result of our routine exercise of plotting frequencies versus ranks. According to the Zipf's law (\citet{piantadosi2014zipf}), the frequency of a word is inversely proportional to its rank. A small proportion of words are used almost all the time, while a vast proportion of words are used rarely.

\begin{figure}[!htb]
\subfigure{\includegraphics[width=0.46\textwidth]{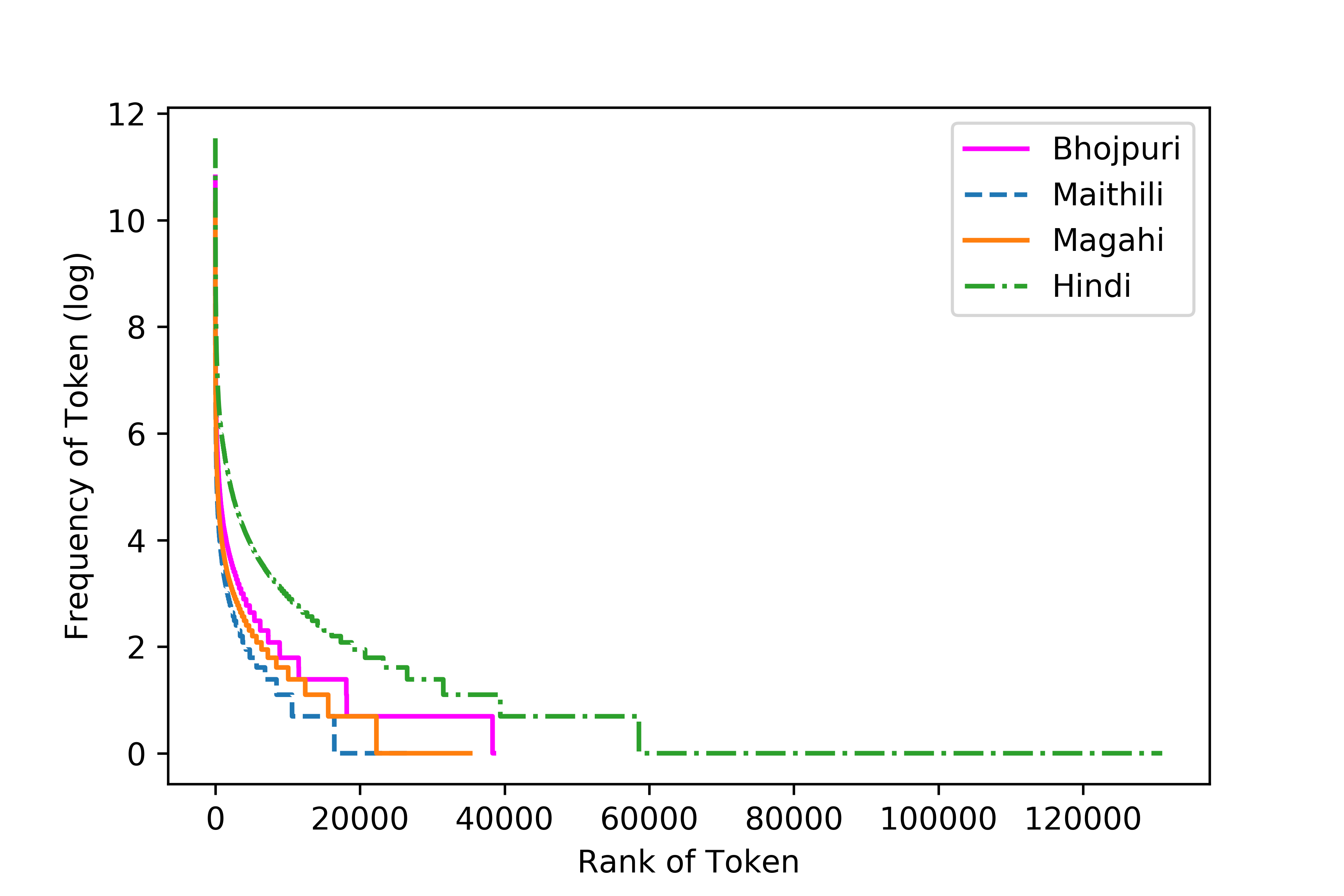}}
\hspace{0.05\textwidth}
\subfigure{\includegraphics[width=0.46\textwidth]{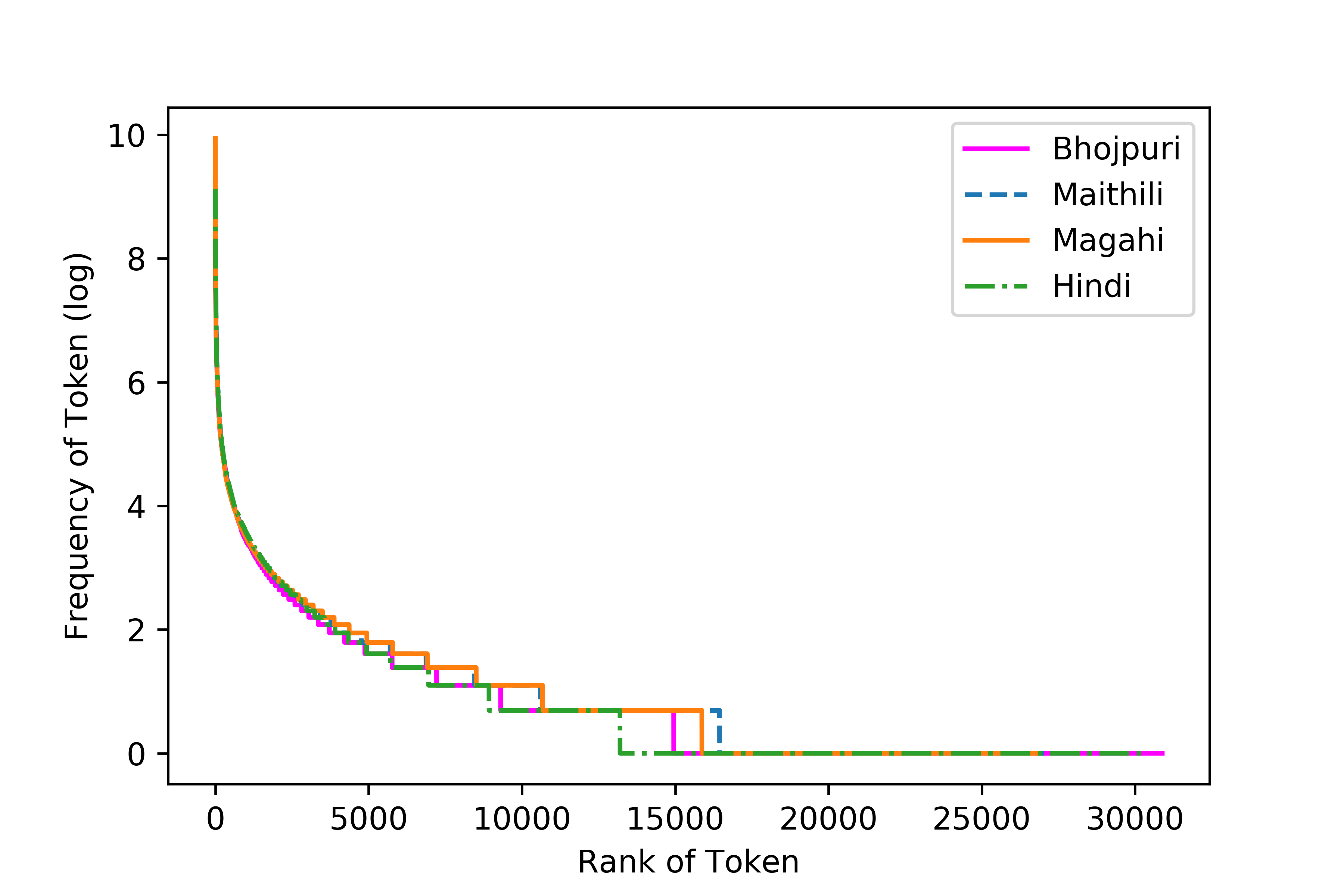}}
\caption{Zipf's curve for Bhojpuri, Maithili, Magahi, and Hindi: The plot on the left is for full corpora, while the one on the right is for minimum (Maithili) size}
\label{fig1}
\end{figure}

The Zipf's curve for Bhojpuri, Maithili, Magahi and Hindi are shown in Figure ~\ref{fig1}. The first Zipf's curve is created by looking at the entire corpus, while the second curve follows the minimum number of sentences (Maithili corpus). In this figure, the frequency is in logarithmic scale. The plots for all the four languages are similar as expected and roughly fit the Zipf's law. It may be noticed that there are far more rare (frequency one) words in Hindi, which is due to the high usage of academic or technical words borrowed from Sanskrit.

\subsubsection{\textbf{Word length}}
\begin{figure}[!ht]
\subfigure{\includegraphics[width=0.46\textwidth]{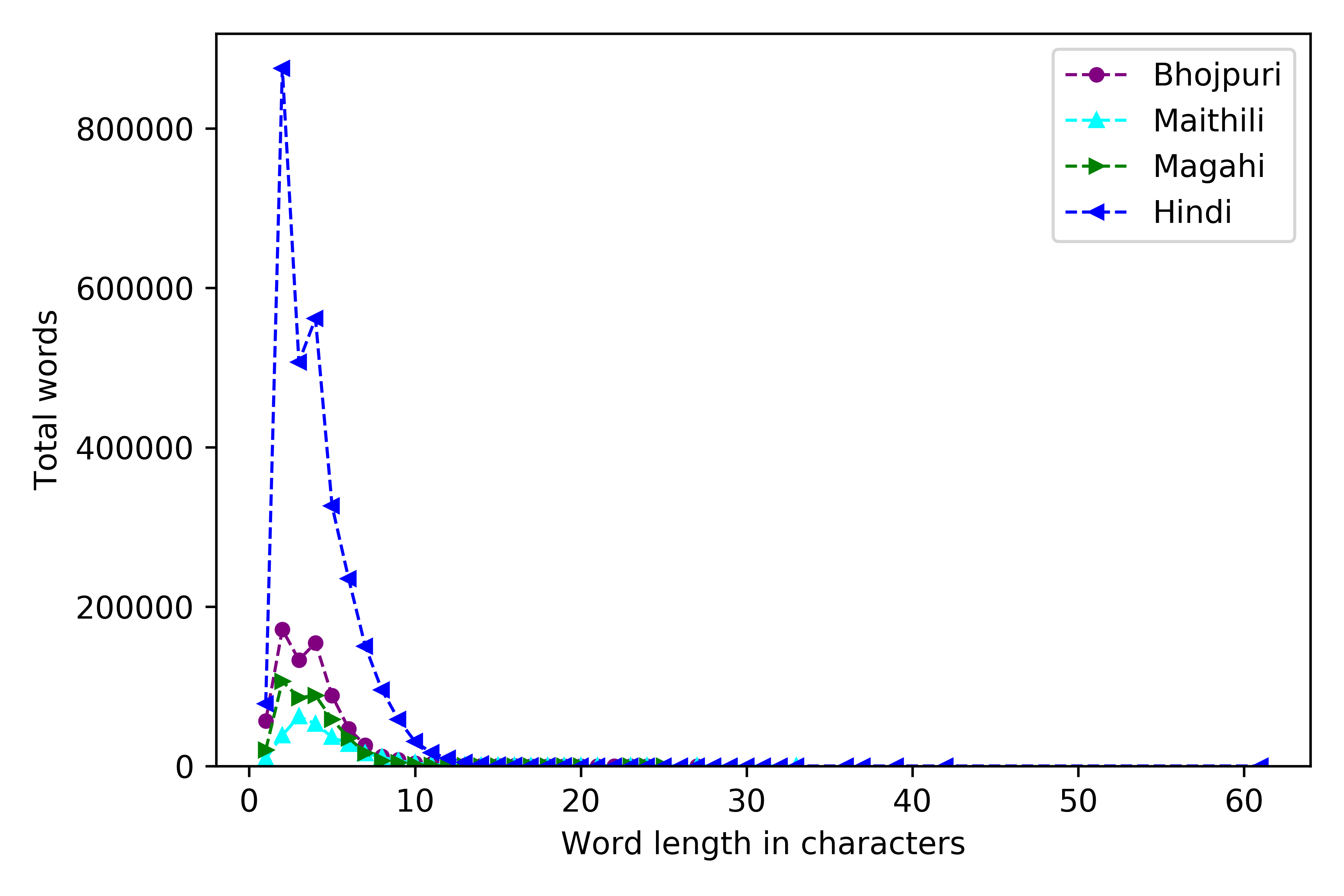}}
\hspace{0.05\textwidth}
\subfigure{\includegraphics[width=0.46\textwidth]{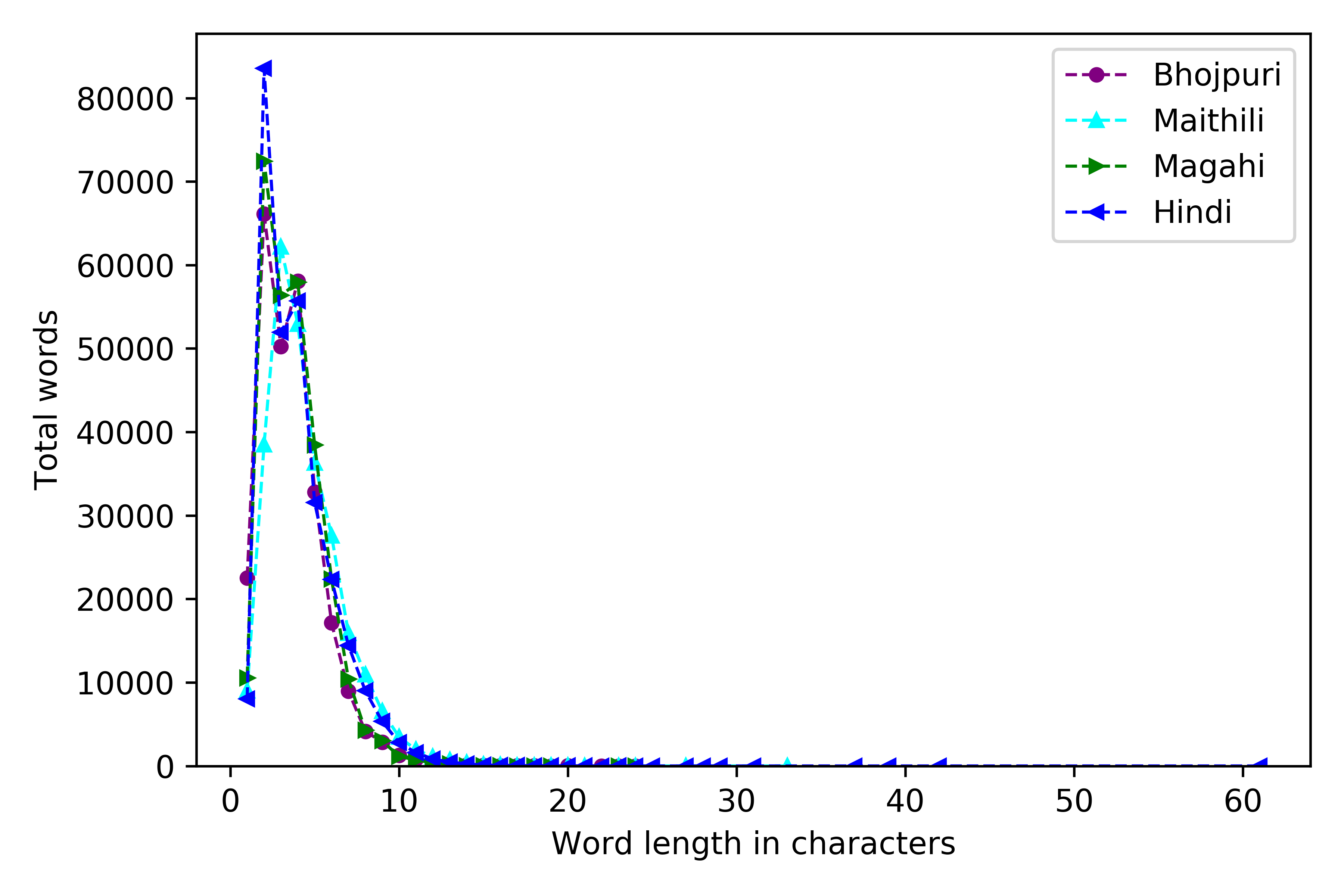}}
\caption{Word length analysis in terms of characters: The plot on the left is for full corpora, while the one on the right is for minimum (Maithili) size}
\label{fig:word_length_analysis}
\end{figure}

Yet another parameter that can be used to compare languages based on basic quantitative measures is the word length. Since the postposition marker or case marker (\emph{vibhakti}) behaves differently in these languages, affecting the word lengths. Sometimes it attaches to the previous word as a morphological suffix, while other times it is a separate word. Word length analysis can give some more evidence in support of or against some existing linguistic knowledge. Also, word length has been an important factor in readability metrics~(\citet{kincaid:75}), although our concern is not about document readability, but language complexity. Table~\ref{length analysis} shows the word length statistics about the four languages and Figure~\ref{fig:word_length_analysis} plots word length versus total words for a certain word length. From the table, the earlier observations about morphological complexity are again strengthened. While the minimum has to be one for all languages, all the other four statistics (maximum, mean, median and standard deviation) have values in the same order as earlier, i.e., highest for Maithili among the three Purvanchal languages and lower for Bhojpuri and Magahi. Hindi has the highest values for the same reasons stated earlier. Bhojpuri, Maithili and Magahi have almost the same metrics scores since morphological complexity is higher than Hindi on the restricted corpus. The figure provides even more evidence for this. It also perhaps shows the typological distance of Maithili from the other two Purvanchal languages, with Hindi being a special case due to relatively high standardization and academic usage.

\begin{table}[!ht]
\caption{Word Length Analysis in terms of characters (where * indicates the number of sentences of each of the languages is made equal to Maithili)}
\label{length analysis}
\centering
\begin{tabular}{cccccccccc}
\toprule
\textbf{Language} &	\textbf{Min} &	\textbf{Max} &	\textbf{Max*} &	
\textbf{Mean} & \textbf{Mean*} &  \textbf{Median} & \textbf{Median*} & \textbf{Std. Dev.} & \textbf{Std. Dev.*} \\
\midrule
Bhojpuri &  1 & 27 & 24 &    13.08 & 12.13 & 13.00 & 12.00 & 7.35 & 6.83 \\
Maithili &  1 & 33 & 33 &    13.52 & 13.52 & 13.00 & 13.00 & 8.15 & 8.15 \\
Magahi &    1 & 25 & 24 &   12.26 & 11.28 & 12.00 & 11.00 & 7.04 & 6.54 \\
Hindi & 	1 & 61 & 61	& 	20.42 & 18.75 & 19.5 & 17.00 & 12.93 & 13.03 \\
\bottomrule
\end{tabular}
\end{table}

On a side note, it may be seen that the word lengths go up to more than 60 characters. This is after removing some `words' from this plot which were not really words but strings left in the corpus due to incorrect tokenization. Even after removing such non-words, on arranging the words in a descending order according to the word length, we found (as expected) that the longest words were either hyphenated words used for literary purposes, or were from quotations from Sanskrit, which is morphologically very complex as well as a fusional language. Non-hyphenated native words in all the four languages rarely exceed the word length of 10, as can be seen from the figure. In the figure, dips and a peaks can be seen. On the left side, the peak is indicative of the corpus size, while on the right (for equal corpus size) it is an indication of the effect of the language family. Dips are present for three of the languages, but not for Maithili. From this figure, the following observations can be made:

\begin{enumerate}
    \item For Hindi, Magahi and Maithili, single character words are mainly or purely punctuation marks (mostly the period symbol). For Bhojpuri, however, there are some single character function words, so that the frequency of single character types is significantly more for Bhojpuri.
    \item For Hindi, Bhojpuri and Magahi, the first and the highest peak occurs at word length two, since two-character function words are the most common and frequent. For Maithili, however, this peak occurs at three, which is in accord with the Table \ref{morpheme frequency}.
    \item The dip, which occurs at word length four for three languages, represents the relationship between (and transition from) the function words and the content words. There is no dip in Maithili, because the transition is smooth, as it has many frequent four character function words as well as content words.
    \item The second peak occurs either because the said transition is not smooth, or because a transition happens suddenly to the most frequent content words at a certain word length.
\end{enumerate}

\subsubsection{\textbf{Entropy and Perplexity}}
Information Theory provides us with some measures to further allow us to compare the complexities and to consider the similarities and the differences of the four languages. Such work has a great deal of precedence, also some specifically for Indian languages (\citet{bharati2000basic}).  

In terms of Information Theory, we can see a sentence as being generated from a probabilistic model such as the $n$-gram model, which is still heavily used, in spite of the availability of deep learning models, which give better performance on many problems. This insight has been used for seemingly very different problems such as protein sequencing or DNA sequencing. In such a view, words are generated one after the other, in sequence, based on their $n$-gram probabilities. At each step of the generation of a sentence, when a word is picked up from the vocabulary, there are as many choices as there are words in the vocabulary.

\begin{table}[!ht]
\caption{Unigram entropy analysis of Purvanchal Languages and Hindi}
\centering
\label{entropy analysis}
\begin{tabular}{ccccc}
\toprule
\textbf{Entropy}	&\textbf{Bhojpuri}	& \textbf{Maithili}	& \textbf{Magahi}  & \textbf{Hindi}\\
\midrule
Complete corpus & 14.48 & 13.47 & 14.08 & 17.44 \\
Restricted corpus & 13.66 & 13.47 & 13.33 & 14.13\\
\bottomrule
\end{tabular}
\end{table}

\begin{table}[!ht]
\caption{Character level entropy analysis of Purvanchal Languages and Hindi}
\centering
\label{char entropy}
\begin{tabular}{ccccc}
\toprule
\textbf{Entropy}	&\textbf{Bhojpuri}	& \textbf{Maithili}	& \textbf{Magahi}  & \textbf{Hindi}\\
\midrule
Complete corpus & 4.86 & 5.01 & 4.96 & 4.98 \\
Restricted corpus & 4.85 & 5.01 & 4.95 & 4.97 \\
\bottomrule
\end{tabular}
\end{table}

Based on this, we get two very basic, but extremely useful measures. One is entropy, which is for the distribution as a whole, while the other is perplexity, which takes into account the sequencing of words in different sentences. Both of them are about the model that we have learnt from the data, but assuming the data is a good representative sample, these measures can give us an indirect estimate of the complexity of the language under consideration. It is important to note here that the complexity we are talking about is at a particular level of linguistic analysis, such as phonological, morphological, syntactic, semantic etc. Here we are mainly concerned with only morphological and syntactic levels. Related to these two levels is the lexical complexity in terms of word types or words in the vocabulary.

\begin{table}[!ht]
\caption{Perplexity of Purvanchal Languages and Hindi (Here * indicates the PP over the equal number of sentences according to Maithili corpus)}
\label{entropy perpexcity}
\centering
\begin{tabular}{ccccccc}
\toprule
\textbf{PP} & \textbf{Unigram} & \textbf{Bigram} & \textbf{Trigram} & \textbf{Unigram*} & \textbf{Bigram*} & \textbf{Trigram*} \\
\midrule
Bhojpuri & 1472.36 & 89.13 & 89.13 & 876.67 & 411.18 & 404.19 \\
Maithili & 1326.84 & 51.11 & 21.29 & 1327.29 & 51.40 & 21.46 \\
Magahi & 762.12 & 222.71 & 201.83 & 814.27 & 166.55 & 140.50 \\
Hindi & 1610.76 & 134.42 & 99.89 & 1140.84 & 662.96 & 660.73 \\
\bottomrule
\end{tabular}
\end{table}

\begin{table}[!ht]
\caption{Character level perplexity of Purvanchal Languages and Hindi (Here * indicates the PP over the equal number of sentences according to Maithili corpus)}
\label{char perplexcity}
\centering
\begin{tabular}{ccccccc}
\toprule
\textbf{PP} & \textbf{Unigram} & \textbf{Bigram} & \textbf{Trigram} & \textbf{Unigram*} & \textbf{Bigram*} & \textbf{Trigram*} \\
\midrule
Bhojpuri & 26.19 & 17.62 & 10.93 & 26.17 & 17.8 & 11.89 \\
Maithili & 25.4 & 15 & 8.75 & 25.4 & 15 & 8.75 \\
Magahi & 26.31 & 19.89 & 16.11 & 27.54 & 21.29 & 17.6 \\
Hindi & 20.38 & 13.03 & 8.91 & 20.69 & 13.97 & 10.61 \\
\bottomrule
\end{tabular}
\end{table}

Table~\ref{entropy analysis} lists the unigram entropy, at word level, of the four languages, which we take as indicative of lexical complexity, i.e., the `richness' of the vocabulary based distribution. Table~\ref{char entropy} gives character unigram entropy calculated over words. It strengthens our previous observations about morphological complexity, with the exception of Hindi. Hindi, again, has morphological complexity mixed up in this metric with the rich vocabulary due to formal use.

Perplexity (PP) of a language is a weighted average of the reciprocal of its branching factor. And the branching factor of a language is the number of possible words that can succeed any word, based on the context as available in the corpus. Therefore, it is a mean representative of the possible succeeding words, given a word. We can roughly consider it as the mean branching factor, and therefore, of the syntactic complexity. If the model is a good enough representation of the true distribution for the language, then the PP value indicates syntactic complexity.

We have used the SRILM toolkit\footnote{\url{http://www.speech.sri.com/projects/srilm/}} for obtaining the $n$-gram based statistical model and for calculating perplexity. We estimated the n-gram language model with backoff smoothing.

From Table~\ref{entropy analysis}, we can see (from the values for full corpus size) that the entropy is the lowest for Maithili and the highest for Hindi. Among the three Purvanchal languages, Maithili is the lowest and Bhojpuri the highest. This again indicates higher morphological complexity for Maithili because it has lower syntactic complexity. The table also indicates high lexical richness for Hindi.

Table~\ref{entropy perpexcity} we observe widely varying values for unigrams, bigrams and trigrams, and also widely varying trends as we go from unigrams to trigrams. We also notice that the values for minimum  corpus size seem to be more meaningful for perplexity comparison than full corpus size, which intuitively makes sense, since the dataset size has a large effect on the perplexity values. Therefore, we focus on the equal corpus size values in this table. From these, we can observe that the value of perplexity for Bhojpuri becomes less than half when we go from unigram to bigram, but decreases only slightly when we go further to trigrams. This may be interpreted to mean that two word groups (or chunks) and three word groups are almost equally syntactically relevant for Bhojpuri. In the case of Magahi, the value drops steeply from unigrams to bigrams and then does not decrease so much. This is similar to the case of Bhojpuri.

We also calculated character level perplexity, that we suggest could be indicative of overall phonological (more accurately, orthographic) and morphosyntactic complexity. If that is the case, which can be further investigated, then it also agrees with the earlier observations. In this context, character level perplexity for words has been related with lexical decision accuracy~(\citet{le-godais-etal-2017-comparing}), although we are not aware of character level perplexity being used for language complexity.

The trend for Hindi is very similar to Bhojpuri, which matches the linguistic intuition that these two languages have the highest similarity, at least as far as is visible in the written corpus available to us.

\section{Experiments on Language Similarity: \emph{SSNGLMScore}}

Preliminary analyses have shown the potential benefits of building tools for resource-poor languages by leveraging resources from resource-rich languages (here, Hindi) based on similarity (\citet{singh2008estimating,wu-etal-2019-language, zoph-etal-2016-transfer, tiedemann-2017-cross}). The similarity among corpora gives a degree of homogeneity among languages (\citet{porta2014using}), motivated by their phylogenetic relationships. Hindi, Bhojpuri, Maithili, and Magahi belong to the Indo-Aryan language family, hence they are phylogenetically related languages, and have typological similarity due to the inheritance of certain features from their common ancestor. We have calculated the character $n$-gram based cross-lingual similarity among these languages. To calculate similarity between two languages, a character-based language model was trained for one language and tested on the other language, and the sentence-wise sum of cross-entropy was obtained. This score should work well as a measure of language similarities, because character level cross entropy is an established measure of language similarity (as used in language identification etc.). It has been suggested previously that many NLP problems can be solved either in part of in full by estimation of the right kind of similarity at the right granularity~\citet{similarity-eurolan:2007,similarity-thesis:2010}. A prominent example of this is language identification. The scaled sum of sentence-wise cross-entropy allows us to compare language similarities on a fair ground. These values are listed in Table~\ref{corpus_similarity}.

\subsection{SSNGLMScore}

We used the KenLm language model tool\footnote{\url{https://kheafield.com/code/kenlm/}}~(\citet{Heafield-estimate}) trained on the corpus using each language ($SL$ $\in$ \{$Bhojpuri$,~$Magahi$,~$Maithili$,~$Hindi$\}) and tested on the corpus of the other three languages $TL$ $\in$ \{$Bhojpuri$,~ $Magahi$,~$Maithili$,~$Hindi$\} with the 5-gram and Kneser-Ney smoothing technique.

Let $m$ sentences exist in a target language ($tl$), and the model was trained on the source language ($sl$), where $sl \in SL$, and $tl \in TL$. The testing scores are scaled in a particular range by the Min-Max scaler algorithm and we then obtained the score by taking the average mean from a language model in the following way:

\begin{equation}
    SS(sl,tl) = \sum_{tl=1}^m p_{sl,tl}(w_n|w_1^{n-1})
\end{equation}

Here SS stands for Similarity Score based on $n$-gram language model scores.

\begin{equation}
    MSS_{sl,tl} = \frac{SS_{sl,tl}-min(SS_{SL,TL})}{max(SS_{SL,TL})-min(SS_{SL,TL})}
\end{equation}

Here MSS stands for Min-Max Scaled Similarity Score.

When the $sl$ and $tl$ both are the same, then the score will be high. For example, when the language model is trained as well as tested on Bhojpuri, it gives a cross-lingual similarity of 1.

Since, to the best of our knowledge, this measure has not been used before for estimating language similarity, we call this measure the Scaled Sum of N-Gram LM Scores or \emph{SSNGLMScore}. We have juxtaposed the obtained similarity scores with the \citet{gamallo2017language} approach of distance scores since both scores provide a -- supposedly inverse -- relationship among languages.

\begin{table}[!ht]
\caption{Character based n-gram corpus similarities based on cross-entropy}
\label{corpus_similarity}
\centering
\begin{tabular}{cc}
\toprule
\textbf{Language Pair}	& \textbf{Cross-entropy score} \\
\midrule
Hindi-Bhojpuri & 3.55\\
Hindi-Maithili & 3.27\\
Hindi-Magahi & 6.04\\
Bhojpuri-Maithili & 7.97\\
Bhojpuri-Magahi & 3.88\\
Maithili-Magahi & 3.68\\
\bottomrule
\end{tabular}
\end{table}

It can be seen from Table~\ref{corpus_similarity} that the linguistic intuition about similarities for these languages is corroborated by the values of cross entropy scores, with a couple of caveats. Maithili is the most distant from Hindi, but is close to Bhojpuri, which matches linguistic knowledge. However, one observable discrepancy is that Bhojpuri is almost equally close to Maithili.

So we look at Table~\ref{corpus_similarity_kenlm} that gives the sequence based similarity scores. These are symmetric scores, as opposed to the cross entropy scores. Bhojpuri is closest to Magahi and almost equally distant from Hindi and Maithili. Maithili is the most distant from Hindi, but almost equally close to the other two Purvanchal languages. Similar reciprocal patterns have been obtained from Table~\ref{corpus_distance_kenlm}, which calculates the distance among languages. 

Magahi is closest to Bhojpuri, then to Maithili and the least to Maithili. This more or less matches the linguistic knowledge. The less expected closeness between Magahi and Hindi is due to the nature of the data as well as the nature of formal written Magahi, which has beeb more heavily infuenced by Hindi, even though it is expected to be more similar to Maithili than Hindi. This is not an uncommon fact for many languages of northern India. For example, formal written Punjabi is much more similar to Hindi, based on the personal knowledge of one of the authors, than the colloquial and rural Punjabi. Some of this may also be affected by chance, because of taking random samples of 50\% corpus size for training and testing. Perhaps $n$-fold cross validation will give us more reliable estimates. It may also indicate that calculating this score at the sentence level is not sufficient.

\begin{table}[!ht]
    \centering
    \caption{Cross-lingual similarity score induced after applying character level language model using SSNGLMScore}
    \begin{tabular}{ccccc}
        \toprule
        \textbf{Language Model} & & \textbf{Test Language} & \\ 
        & \textbf{Bhojpuri} & \textbf{Hindi} & \textbf{Magahi} & \textbf{Maithili} \\ 
        \midrule
        Bhojpuri & 1 & 0.2398 & 0.3177 & 0.1331 \\ 
        Hindi &  & 0.8676 & 0.2120 & 0.0874 \\ 
        Magahi &  &  & 0.9863 & 0.1173 \\ 
        Maithili &  &  &  & 0.9543 \\ 
        \bottomrule
    \end{tabular}
    \label{corpus_similarity_kenlm}
\end{table}{}

\begin{table}[!ht]
    \centering
    \caption{Cross-lingual distance score induced after applying character level language model using perplexity based score (\citet{gamallo2017language})}
    \begin{tabular}{ccccc}
        \toprule
        \textbf{Language Model} & & \textbf{Test Language} & \\ 
        & \textbf{Bhojpuri} & \textbf{Hindi} & \textbf{Magahi} & \textbf{Maithili} \\ 
        \midrule
        Bhojpuri & 0 & 0.8571 & 0.4245 & 0.6015 \\ 
        Hindi &  & 0 & 0.8164 & 0.7850 \\ 
        Magahi &  &  & 0 & 0.6825 \\ 
        Maithili &  &  &  & 0 \\ 
        \bottomrule
    \end{tabular}
    \label{corpus_distance_kenlm}
\end{table}{}

From Table~\ref{corpus_similarity_kenlm} and Table~\ref{corpus_distance_kenlm} have been observed that the Bhojpuri and Magahi are similar language and also have minimum distance than others.

\section{Bilingual Dictionary and Synset}
Bilingual lexicon and WordNet~(\citet{miller1990}) are crucial linguistic resources for various NLP applications, such as Machine Translation  (\citet{madankar2016information, gracia2019results, zhang2019improving}), Cognate Identification (\citet{wang2014multilingual,nakov2009unsupervised, ciobanu-dinu-2013-dictionary,nasution2016constraint}), Part-of-Speech Tagging (\citet{fang-cohn-2017-model, prabhu2019extraction}), Word-sense Disambiguation (\citet{vaishnav2019knowledge,jain2019word}), Bilingual Lexicon Induction (\citet{zhang2017bilingual,hangya2018two, nasution2016constraint}), Word Alignment (\citet{probst2002using}), Cross-language Information Retrieval (\citet{sharma2018cross, madankar2016information}), Word Translation (\citet{lample2018word}), and Cross-lingual Word-embedding Alignment (\citet{ duong-etal-2016-learning,ruder2019survey}).

Since this work was directed towards developing Machine Translation systems between Hindi and the three Purvanchal languages, we had undertaken to initiate building WordNets for these languages as well. Since there is already a linked WordNet for many Indian languages, which is called IndoWordNet~(\citet{Bhattacharyya10indowordnet}), in which WordNets of different languages are linked together by mapping synsets (sets of synonyms) of different languages to those of Hindi, i.e., Hindi is used as the Inter-Lingual Index. However, as part of our project, we only created the synsets and mapped them to Hindi. Bilingual lexicon for the three languages are prepared in the usual way, as suitable for machine translation.

The information in a synset includes IDs (corresponding to word meanings in Hindi as given in IndoWordNet), synonyms, lexical categories, and concepts for each entry. The Hindi synset\footnote{\url{http://www.cfilt.iitb.ac.in/wordnet/webhwn/}} and bilingual dictionary (Shabdanjali\footnote{\url{http://ltrc.iiit.ac.in/onlineServices/Dictionaries/Dict_Frame.html}}, a English-Hindi open source e-dictionary) have been used for comparison. The sizes of these two resources are given in Table \ref{bi_dict_and_synset}.

\begin{table}[!ht]
\caption{Cumulative resource size of Bilingual dictionary (Word-meanings) and Synsets}
\label{bi_dict_and_synset}
\centering
\begin{tabular}{ccccc}
\toprule

\textbf{Resources}	& \textbf{Bhojpuri} & \textbf{Maithili} & \textbf{Magahi} & \textbf{Hindi}\\

\midrule
Bilingual dictionary & 12211 & 7485 & 15379 & 27007\\
Synset & 37421 & 8217 & 10439 & 39715\\

\bottomrule
\end{tabular}
\end{table}

The created synset has the highest number of total synonyms and synonyms per word for Bhojpuri. Although Magahi has more cumulative synonyms, Maithili closely surpasses Magahi in synonyms per word. The synonym-statistics is given in Table ~\ref{synset_statics}. Each entry in this table correspond to the total number of synonyms per word. For instance, entry in column 1 signify that minimum 1 synonym per word is present in the synset. Similarly, columns 2, 3 and 4 indicate the maximum, and mean of synonyms per word in the corresponding language. It is not clear if any conclusions can be drawn from these statistics alone. However, the data could be useful for studying, say, the degree of ambiguity in these languages.\\

\begin{table}[!ht]
\caption{Distribution of synonyms per word in the created synset for the three languages, compare with Hindi synset}
\label{synset_statics}
\centering
\begin{tabular}{cccc}
\toprule

\textbf{Language}	& \textbf{Min} & \textbf{Max} & \textbf{Mean} \\

\midrule
Bhojpuri & 1 & 101 & 2.93 \\
Magahi & 1 & 95 & 2.35  \\
Maithili & 1 & 41 & 2.50 \\
Hindi & 0 & 140 & 2.75 \\
\bottomrule
\end{tabular}
\end{table}

\section{Shallow Syntactic Annotation}
Annotation enriches the raw corpus with information and allows both linguistic analysis and machine learning. However, annotation by a human is not guaranteed to be perfect. It has been shown that annotation can reflect the annotator's errors and predilections (\citet{wynne2005developing}). Therefore, we try to discuss some of the annotation issues that we faced while preparing the annotated corpus.

Syntactic annotation can be very useful for numerous problems in NLP, whether it is Machine Translation, Automatic Speech Recognition or Information Retrieval. Since this work can be seen as a continuation of the ILMT project (Section 2), we used the standard BIS (Bureau of Indian Standards) tagset (\citet{bharati2006anncorra}) for Indian languages that was used in the ILMT project. The BIS tagset consists of 25 POS tags (Table~ \ref{pos tagset}) and 11 chunk tags\footnote{The chunk annotation was performed for Bhojpuri and Maithili languages only} (Table~ \ref{tab:chunk_tagset}) for Bhojpuri,  Maithili, and Magahi language. 
This tagset was used with the understanding that if it can be used for most major Indian languages, it can also be used for the Purvanchal languages. However, for modeling language-specific properties, an extension of this tagset may be required since, according to some researchers, it is not suitable as it is for all Indian languages (\citet{kumar2012developing}).

Since we did not go beyond Chunking, our syntactic annotation is basically shallow syntactic annotation, which can be useful for building, for example, shallow parsers for these languages (\citet{Sha2003shallowparsing,Sangal2007shallowparsing}), or for shallow parsing based machine translation systems.

\begin{table}[!ht]
    \caption{BIS tagset used in POS annotation}
    \label{pos tagset}
    \centering
    \begin{tabular}{ l l l |l l l}
    \toprule
         \textbf {S.No.} & \textbf {Tag} & \textbf {Description} & \textbf {S. No.} & \textbf {Tag} & \textbf{Description} \\
    \midrule
        1 & NN & NOUN & 14 & QO & ORDINAL \\
        2 & NNP & PROPER NOUN & 15 & SYM & SPECIAL SYMBOL \\
        3 & PRP & PRONOUN & 16 & CC & CONJUNCTS \\
        4 & VM & VERB MAIN & 17 & CL & CLASSIFIER \\
        5 & VAUX & VERB AUXILIARY & 18 & QF & QUANTIFIER \\
        6 & DEM & DEMONSTRATIVE & 19 & INJ & INTERJECTION \\
        7 & PSP & POST POSITION & 20 & INTF & INTENSIFIER \\
        8 & QC & CARDINAL & 21 & NEG & NEGATIVE \\
        9 & JJ & ADJECTIVE & 22 & RDP & REDULICATIVE \\
        10 & RB & ADVERB & 23 & ECH & ECHO \\
        11 & NST & NOUN LOCATION & 24 & UNK & UNKNOWN \\
        12 & WQ & QUESTION WORD & 25 & UT & QUOTATIVE\\
        13 & RP & PARTICLE   \\
    \bottomrule
    \end{tabular}
\end{table}{}

\begin{table}[!ht]
    \caption{BIS tagset used in Chunk annotation}
    \label{tab:chunk_tagset}
    \centering
    \begin{tabular}{lll}
    \toprule
        \textbf{S. No.} & \textbf{Tag} & \textbf{Description} \\
    \midrule
        1 & NP & Noun Chunk \\
        2 & VGF & Finite Verb Chunk \\
        3 &  VGNF & Non-finite Verb Chunk \\
        4 & VGINF & Infinitival Verb Chunk \\
        5 & VGNN & Gerunds \\
        6 & JJP & Adjectival Chunk \\
        7 & RBP & Adverb Chunk \\
        8 & NEGP & Negatives \\
        9 & CCP & Conjuncts \\
        10 & FRAGP & Chunk Fragments \\
        11 & BLK & Miscellaneous entities \\
    \bottomrule
    \end{tabular}
\end{table}

Although it is possible to perform annotation in a simple text editor, since the annotated file is stored in a text file, such an approach can lead to easily avoidable errors such as appending operating system specific encoding characters, junk characters, or due to typos. It is, therefore, a common practice to use an annotation tool with a Graphical User Interface that supports a degree of error checking. We used the same tool for such annotation as was used for the ILMT project, namely Sanchay (\citet{singh2008mechanism}). As far as actual file format or representation is concerned, we used the Shakti Standard Format (SSF), which is a robust recursive representation. It allows storing a wide variety of linguistic and non-linguistic information in one place. The detailed description of the SSF representation can be found in \citet{bharati2014ssf}. We tried to get the same kind of statistics as above for POS tags, pairs of words and POS tags and of chunks. Table~\ref{pos and chunk entropy},~\ref{pos and chunk mattr} show the TTR, entropy and MATTR values for these units, respectively.

Entropy values do not vary much for the three units over the four languages. Only two things are noticeable here from the second column of the Tables ~\ref{pos and chunk entropy} and ~\ref{pos and chunk mattr}. One is that Word\_POS\_Tag TTR and MATTR are quite low for Hindi and, for the same language, the Chunk Tag TTR and MATTR are very high for Hindi, while POS Tag TTR and MATTR are almost the same for all languages. This seems to indicate that syntactic complexity of the three languages in terms of POS tags is almost the same. The Word\_POS\_Tag TTR is twice for Hindi that the other three languages, meaning lexicalized POS tag complexity (say, as modelled by HMM) is higher for Hindi. Chunk TTR and MATTR being very high for Hindi indicates much greater syntactic complexity at the chunk tag level, which could partially be explained by the same reason of academic usage, but also of larger data size. Hindi is also relatively less morphologically complex, which could suggest it has relatively higher syntactic complexity. We do not rely as much on varied data for the three Purvanchal languages as we have for Hindi. This factor might be affecting other measures too.

\begin{table}[!ht]
    \caption{Entropy and TTR calculations for POS Tag, Word\_POS\_Tag pair, and Chunk Tag annotation}
    \label{pos and chunk entropy}
    \centering
    \begin{tabular}{cccc|ccc}
    \toprule
        & & \textbf{Entropy} & & & \textbf{TTR} & \\
    \toprule
        \textbf{Languages} & \textbf{POS Tag} & \textbf{Word\_POS\_Tag} & \textbf{Chunk Tag}  & \textbf{POS Tag} & \textbf{Word\_POS\_Tag} & \textbf{Chunk Tag} \\
    \midrule
        Bhojpuri & 3.35 & 10.93 & 2.60 & 0.0106 & 13.49 & 0.0319\\
        Maithili & 3.55 & 11.27& 3.44 & 0.0130 & 13.11 & 0.1231\\
        Magahi & 3.44 &  10.62 & - & 0.0152 & 11.97 & -\\
        Hindi & 3.63 & 10.38 & 2.62 & 0.0135 & 6.50 & 7.32\\
    \bottomrule
    \end{tabular}
\end{table}

\begin{table}[!ht]
    \caption{MATTR calculations for POS Tag, Word\_POS\_Tag pair, and Chunk Tag annotation}
    \label{pos and chunk mattr}
    \centering
    \begin{tabular}{cccc}
    \toprule
        \textbf{Languages} & \textbf{POS Tag} & \textbf{Word\_POS\_Tag} & \textbf{Chunk Tag} \\
    \midrule
        Bhojpuri & 0.037 & 0.580 & 0.019 \\
        Maithili & 0.039 & 0.595 & 0.022 \\
        Magahi & 0.037 & 0.591 & -\\
        Hindi & 0.044 & 0.523 & 0.022\\
    \bottomrule
    \end{tabular}
\end{table}

\subsection{Annotation Issues}
There are many challenges while annotating the corpora of these less resourced languages. Bhojpuri, Maithili and Magahi are partially synthetic languages because they often express syntactic relations within word forms through suffixes, as shown in some of the examples given in Section \ref{bmm-differences}. Thus, the use of embedded case markers, emphatic markers, classifiers, determiners etc. is frequent in these languages. This is responsible for many problems in annotation.

\subsubsection{\bf Embedded Case Markers}
It is already mentioned in Section \ref{bmm-differences} that in many morphological constructions of these languages, case markers get merged with nominals and pronominals, especially locative, instrumental and genitive case markers get fused with nominals and pronominals.

The annotated examples listed here are the problematic cases because the presently assigned tags are unable to capture the full meaning of such nominal and pronominal fused forms. This can be a matter for further work on these languages. Perhaps the tagset also may have to be revised a little to accommodate certain idiosyncrasies of these languages. 

\subsubsection{\bf Diverse Realizations of Single Token}
In Bhojpuri, Maithili and Magahi, several cases of diverse realization of a single token are found. The token [ke] has the highest different realizations in Magahi and Bhojpuri: 12 tags and 15 tags respectively. Whereas in Maithili the token [je] has been assigned 14 tags for different realizations. List of such tokens is long in all these three languages. Mostly these tokens have multiple functional and connotative meanings, making the process of annotation hard and error prone. 

\subsubsection{\bf Homophones}
Like many other Indian languages, Bhojpuri, Maithili and Magahi also have homophone words in abundance. These words look similar but their POS tags are varied which may pose a sense of perplexity for the annotator. 
 \newline 
Example-1: \\
\tab \textit{Bhojpuri}: [kula] (JJ) paisave (NN) orA (VM) gaila (VAUX) . (SYM) \\ 
\tab \textit{Hindi}: [sArA] pEsA hI samApwa ho gayA. \\ 
\tab \textit{English Translation}: [All] the money is gone. 
\\
\tab \textit{Bhojpuri}: wohAra (PRP) kOna (WQ) [kula] (NN) bA (VM) ? (SYM) \\
\tab \textit{Hindi}: wumhArA kOna sA [kula] hE? \\
\tab \textit{English Translation}: Which one is your [clan]? \\
 In these example sentences of Bhojpuri, the word [kula] is homophones, each has a different meaning and different grammatical category. \\
\\
Example-2: \\
\tab \textit{Bhojpuri}: laikA (NN) geMxa (NN) ke (PSP) [loka] (VM) lihalasa (VAUX) . (SYM) \\
\tab \textit{Hindi}: ladZake ne geMxa [lapaka] lI. \\
\tab \textit{English Translation}: Boy [caught] the ball.
\\
\tab \textit{Bhojpuri}: sahara (NN) A (CC) gazvaI (JJ) [loka] (NN) meM (PSP) aMwara (NN) bA (VM) . (SYM) \\
\tab \textit{Hindi}: Sahara Ora grAmINa [samAja] meM aMwara hE. \\
\tab \textit{English Translation}: There is difference between the urban and the rural [society]. \\
In these example sentences of Bhojpuri, the word [loka] is homophonous,  with each homophone having a different meaning and different grammatical category. 
\\

\subsubsection{\bf Amalgamated Emphatic Expressions}
In these languages, emphatic particles are generally merged with nominals and pronominals and hence they make annotation task difficult, as mentioned in Section \ref{bmm-differences}.

\subsection{Annotated Corpus Statistics}

Earlier reported work on Bhojpuri and Magahi had 120k, 63k tokens, respectively, annotated with modified tagset of Bureau of Indian Standards or BIS  (\citet{ojha2015training, kumar2012developing}). Bhojpuri, Maithili, and Magahi being under-resourced languages, researchers have not paid that much attention to these languages, even though a large number of native speakers are available. One earlier work on Maithili POS tool is available online\footnote{\url{http://sanskrit.jnu.ac.in/mpost/index.jsp}}. Recently, \citet{priyadarshi2020towards} have reported a POS tagger and studied the influence of the language-specific features (\citet{priyadarshi2020linguistic_features}).

In our work on annotation, the cleaned raw corpus was used for POS tag and Chunk annotation. POS annotated corpus comprises of 245489 words for Bhojpuri, 208648 words for Maithili and 171538 million words for Magahi. Also, Chunk annotation has 9695 and 1954 sentences (60591 and 10476 tokens, respectively) for Bhojpuri and Maithili. For Hindi, we used the Hindi-Urdu dependency treebank dataset, which comprises 20866 sentences, which have 436940 POS tags and 235426 chunks. The annotation statistics of these language given in Table~ \ref{pos annotation}.

\begin{table}[!ht]
\caption{Annotation statistics}
\centering
\label{pos annotation}
\begin{tabular}{ccccc}
\toprule
\textbf{Language} & \multicolumn{2}{c}{\textbf{POS annotation}} & \multicolumn{2}{c}{\textbf{Chunk annotation}} \\
 & \textbf{\# Sentence}   & \textbf{\# Token}   & \textbf{\# Sentence}  & \textbf{\# Token} \\ \midrule

Bhojpuri & 16067  & 245482 & 9695 & 60588 \\ 
Maithili & 12310 & 208640 & 1954 & 10476 \\ 
Magahi  & 14669 & 171509 & - & -\\ \bottomrule
\end{tabular}
\end{table}

The POS tag distribution in the annotated corpus is given in Table ~\ref{tab:tagtable}. The NN, VM, PSP tags covers approximately 50\% of Bhojpuri annotated corpus, whereas the same is true for NN, VM, PSP, SYM tags in Maithili and NN, VM, SYM tags in Magahi annotated corpus include. Moreover, some tags having a coverage of less than 0.1\% are considered as rare tags, such as, ECH, UNK, UT for Bhojpuri, ECH, UT for Maithili and CL, UNK, UT for Magahi, These have nearly 0\% coverage of POS tag distribution in annotated corpus.

In the chunk tag annotated corpus, the Noun Phrase (NP) is the most frequent as compared to other phrases in Bhojpuri and Maithili. Moreover, Chunk Fragment Phrase (FRAGP) has the minimum occurrence, as mentioned in Table ~\ref{tab:chunktable}.

\begin{table}[!ht]
\tiny
\caption{Frequency based POS tag percentage ratio of Bhojpuri, Maithili, Magahi and Hindi annotated corpus. Here only those POS tags of Hindi are considered, which were used for annotation of the three languages}
\centering
\label{tab:tagtable}
\begin{tabular}{|c|c|c|c|c|c|c|c|c|}
\hline
\textbf{POS Tag} & \textbf{Tag Frequency} & \textbf{Tag Percentage} & \textbf{Tag Frequency} & \textbf{Tag Percentage} & \textbf{Tag Frequency} & \textbf{Tag Percentage} & \textbf{Tag Frequency} & \textbf{Tag Percentage}\\
& (Bhojpuri) & (Bhojpuri) & (Maithili) & (Maithili) & (Magahi) & (Magahi) & (Hindi) & (Hindi)\\ \hline
        
        NN  & 64659 & 26.34 & 54971 & 26.35 & 42925 & 25.03 & 86742 & 19.85 \\ \hline 
        VM  & 38098 & 15.52 & 26226 & 12.57 & 21678 & 12.64 & 46500 & 10.64 \\ \hline 
        PSP  & 35226 & 14.35 & 17012 & 8.15 & 17933 & 10.46 & 83879 & 19.19 \\ \hline 
        SYM  & 28993 & 11.81 & 21108 & 10.12 & 26167 & 15.26 & 31445 & 7.1 \\ \hline 
        PRP  & 16655 & 6.78 & 12976 & 6.22 & 5840 & 3.41 & 18486 & 4.23 \\ \hline 
        VAUX  & 10740 & 4.38 & 18028 & 8.64 & 17094 & 9.97 & 28195 & 6.45 \\ \hline 
        NNP  & 9232 & 3.76 & 13381 & 6.41 & 4111 & 2.4 & 33519 & 7.67 \\ \hline 
        JJ  & 8781 & 3.58 & 9124 & 4.37 & 6053 & 3.53 & 23380 & 5.35 \\ \hline 
        CC  & 7476 & 3.05 & 7600 & 3.64 & 4747 & 2.77 & 16070 & 3.67 \\ \hline 
        RP  & 4287 & 1.75 & 3076 & 1.47 & 5247 & 3.06 & 7233 & 1.65 \\ \hline 
        NST  & 4092 & 1.67 & 5148 & 2.47 & 4694 & 2.74 & 6488 & 1.48 \\ \hline 
        QF  & 3590 & 1.46 & 3560 & 1.71 & 2480 & 1.45 & 3425 & 0.78 \\ \hline 
        QC  & 3385 & 1.38 & 4324 & 2.07 & 1780 & 1.04 & 7396 & 1.69 \\ \hline 
        NEG  & 3197 & 1.3 & 2232 & 1.07 & 2140 & 1.25 & 3253 & 0.74 \\ \hline 
        WQ  & 1811 & 0.74 & 1770 & 0.85 & 1221 & 0.71 & 724 & 0.16 \\ \hline 
        DEM  & 1429 & 0.58 & 2761 & 1.32 & 2122 & 1.24 & 5642 & 1.29 \\ \hline 
        RB  & 1239 & 0.5 & 2917 & 1.4 & 3500 & 2.04 & 1956 & 0.44 \\ \hline 
        RDP  & 722 & 0.29 & 320 & 0.15 & 205 & 0.12 & 365 & 0.08 \\ \hline 
        QO  & 513 & 0.21 & 654 & 0.31 & 571 & 0.33 & 742 & 0.16 \\ \hline 
        INTF  & 499 & 0.2 & 295 & 0.14 & 489 & 0.29 & 438 & 0.1 \\ \hline 
        INJ  & 418 & 0.17 & 413 & 0.2 & 288 & 0.17 & 164 & 0.03 \\ \hline 
        CL  & 248 & 0.1 & 295 & 0.14 & 66 & 0.04 & 0  & 0 \\ \hline 
        ECH  & 122 & 0.05 & 54 & 0.03 & 129 & 0.08 & 1 & 0 \\ \hline 
        UNK  & 63 & 0.03 & 394 & 0.19 & 28 & 0.02 & 205 & 0.04 \\ \hline 
        UT  & 7 & 0 & 1 & 0 & 1 & 0 & 0 & 0 \\ \hline 
\end{tabular}
\end{table}

\begin{table}[!ht]
\small
\caption{Frequency based Chunk tag percentage ratio of Bhojpuri, Maithili and Hindi annotated corpus. Here only those Chunk tags of Hindi are considered, which were used for annotation of the three languages}
\centering
\label{tab:chunktable}
\begin{tabular}{|c|c|c|c|c|c|c|}
\hline

\textbf{Chunk Tag} & \textbf{Tag Frequency} & \textbf{Tag Percentage} & \textbf{Tag Frequency} & \textbf{Tag Percentage} & \textbf{Tag Frequency} & \textbf{Tag Percentage}\\
& (Bhojpuri) & (Bhojpuri) & (Maithili) & (Maithili) & (Hindi) & (Hindi)\\ \hline

NP  & 35393 & 58.42 & 4321 & 41.25 & 135423 & 57.52 \\ \hline 
VGF  & 16951 & 27.98 & 3069 & 29.3 & 32886 & 13.96 \\ \hline 
CCP  & 3394 & 5.6 & 1125 & 10.74 & 16119 & 6.84 \\ \hline 
VGNF  & 1669 & 2.75 & 49 & 0.47 & 5250 & 2.23 \\ \hline 
BLK  & 1559 & 2.57 & 47 & 0.45 & 21828 & 9.27 \\ \hline 
JJP  & 690 & 1.14 & 555 & 5.3 & 10309 & 4.37 \\ \hline 
NEGP  & 447 & 0.74 & 347 & 3.31 & 379 & 0.16 \\ \hline 
RBP  & 397 & 0.66 & 458 & 4.37 & 2268 & 0.96 \\ \hline 
VGNN  & 54 & 0.09 & 395 & 3.77 & 8344 & 3.54 \\ \hline 
VGINF  & 32 & 0.05 & 108 & 1.03 & 0 & 0 \\ \hline 
FRAGP  & 2 & 0 & 2 & 0.02 & 412 & 0.17 \\ \hline 
\end{tabular}
\end{table}

We calculated the inter-annotator agreement among the random annotated sets of 962 for Bhojpuri, 578 for Magahi and 381 for Maithili using Cohen’s Kappa (\citet{doi:10.1177/001316446002000104}) to validate the quality of annotation. Cohen’s Kappa inter-annotator score was 0.92, 0.64, and 0.74 for Bhojpuri, Magahi and Maithili, respectively, which indicates that the quality of the annotation and presented schema is usable to different degrees.

\subsection{Baseline for POS Tagging and Chunking for Purvanchal Languages}

Since the objective of this work as to function as the basis for further work on these languages, we finish the section on annotation by describing a reasonably good baseline, which can be treated as the lower baseline for later research on building POS taggers and Chunkers of these languages.

As part of our experiment with this baseline, we compare our results with those for Hindi part of the Hindi-Urdu  dependency treebank for automatic annotation. Since we suspect the quality of the data and annotation is not as good as, say, for Hindi, we also consider these experiments as a kind of additional validation that our data may be useful for the purposes for which it was made. The idea is that if a reasonably good machine learning algorithm is able to give tolerable predictions (which could be improved in future) using this data, then the data is not useless, in spite of shortcomings.

For the experiments, we have used the traditional state-of-the-art machine learning or probabilistic technique, known as the Conditional Random Fields (CRF), for sequence labeling. The CRF model provides surprisingly good results, even in the absence of an ample amount of annotated dataset. CRF++ toolkit{\footnote{https://taku910.github.io/crfpp/}} with the default feature set and parameters have been used for POS tagging and Chunking. For performing these experiments, we have divided the dataset into 80-20 ratio for training and testing. The reported results for POS tagging and Chunking of each language are evaluated on the testing dataset. The feature-wise obtained results for POS tagging and Chunking in terms of weighted F$_1$-score are shown in Table~\ref{tab:feat_pos} and ~\ref{tab:feat_chunk}, respectively. We have used the previous word, next word and contextual word as a feature for POS tagging. Additionally, the contextual POS tag is also considered as a feature for Chunking. The contextual information (both word and POS tag) play an essential role in improving model performance, as shown in these tables.  

\begin{table}[!h]
    \caption{Effect of features in POS tagging's F$_1$-score in \%}
    \label{tab:feat_pos}
    \centering
    \begin{tabular}{l|cccc}
    \toprule
        \textbf{Features} & 	\textbf{Bhojpuri} &	\textbf{Maithili} &	\textbf{Magahi} &	\textbf{Hindi} \\
    \midrule
        Word	& 0.80 &	0.72 &	0.76 &	0.85 \\
        +Previous words	& 0.82	& 0.75	& 0.77	& 0.89 \\
        +Next words & 	0.85	& 0.75	& 0.78	& 0.93 \\
        +Group words	& 0.88	& 0.76	& 0.78	& 0.94 \\
    \bottomrule
    \end{tabular}
\end{table}

\begin{table}[!h]
    \caption{Effect of features in Chunking's F$_1$-score in \%}
    \label{tab:feat_chunk}
    \centering
    \begin{tabular}{l|ccc}
    \toprule
        \textbf{Features} & 	\textbf{Bhojpuri} &	\textbf{Maithili} &	\textbf{Hindi} \\
    \midrule
        Word &	0.79 &	0.64 & 0.92 \\
        +Previous words &	0.83 &	0.66 &	0.94 \\
        +Next words &	0.91 &	0.93 &	0.98 \\
        +Contextual tags &	0.92 &	0.93 &	0.97 \\
        +Group words &	0.92 &	0.94  & 0.99 \\
    \bottomrule
    \end{tabular}
\end{table}

Furthermore, the obtained results are compared with the Hindi dependency treebank dataset in terms of accuracy, weighted precision, weighted recall, and weighted F$_1$-score, which are shown in the Tables ~\ref{pos_result} and \ref{chunk_result}.

For POS tagging, we get 88\% F$_1$-score Bhojpuri, which is not bad considering that we used the default template. For Maithili and Magahi, the values are a bit below the 80\% mark, which can be explained by the low quantity of data. Compared to Hindi, the results are not very unexpected.

For Chunking, the results for Bhojpuri and Maithili are even better, being above 90\%, although much lower than the 99\% for Hindi. These values are higher for Chunking than for tagging because the number of Chunk tags is less than half of the number of POS tags. Moreover, only two Chunk tags (noun phrase and finite verb phrase) cover most of the corpus for all three languages. Also, due to their higher morphological complexity, what would have been a multi-word chunk often becomes a single word chunk in these languages, leading to easier prediction of chunks.

\begin{table}[!h]
    \caption{POS tagging experimental result in \% using CRF technique}
    \centering
    \label{pos_result}
    \begin{tabular}{ccccc}
         \toprule
         \textbf{Language} & \textbf{Precision} & \textbf{Recall} & \textbf{F$_1$-score} & \textbf{Accuracy} \\
         \midrule
        \textbf{Bhojpuri} & 0.89 & 0.89 & 0.88 & 0.89 \\
        \textbf{Maithili} & 0.77 & 0.77 & 0.76 & 0.77 \\
        \textbf{Magahi} & 0.8 & 0.78 & 0.78 & 0.78 \\ 
        \textbf{Hindi} & 0.94 & 0.94 & 0.94 & 0.94 \\ 
        \bottomrule
    \end{tabular}
\end{table}

\begin{table}[!h]
    \caption{Chunking experimental result in \% using CRF technique}
    \centering
    \label{chunk_result}
    \begin{tabular}{ccccc}
         \toprule
         \textbf{Language} & \textbf{Precision} & \textbf{Recall} & \textbf{F$_1$-score} & \textbf{Accuracy} \\
         \midrule
            \textbf{Bhojpuri} & 0.92 & 0.92 & 0.92 & 0.92 \\
            \textbf{Maithili} & 0.94 & 0.94 & 0.94 & 0.94 \\
            \textbf{Hindi} & 0.99 & 0.99 & 0.99 & 0.99 \\
        \bottomrule
    \end{tabular}
\end{table}

\section{An Experiment on Language Identification}
 
Language Identification (LI) on text is one of the most basic tasks in NLP, which determines the language of the provided text, where the input text is either a word, a sentence, paragraph or even a document or a set of documents. 
LI aims to mimic human ability to recognise or identify a language, without actually knowing that language. Multiple approaches have been proposed over the years, that infer the languages without human intervention. According to \citet{simons2017ethnologue}, LI approaches should able to discern thousands of human language.

LI is one of the oldest problems in NLP. The text displayed on computing devices is made up of a stream of characters (or glyphs), which are digitised by a particular encoding scheme. Most troubling for LI is the use of either isomorphic or proprietary encodings to encode text in some languages where there is a lack of standardization or standardization has not yet been fully adapted.

Perhaps the most important advance was the \citet{cavnar1994n} method, which has popularised the use of character level n-gram models for automatic LI.

It is useful in various NLP applications which include authorship profiling, machine translation, information retrieval, lexicography and many more. Moreover, the LI system's output is utilized in the adaptation of boilerplate NLP tools that require annotated data, such as POS tagger, Chunker for resource-poor languages.

Earlier work considers encoding in terms of bytes (\citet{singh2006study,singh2007identification}), character (\citet{singh2006study,singh2007identification,baldwin2010language,damashek1995gauging}), character combination (\citet{windisch2005language,sterneberg2012language,banerjee2014hybrid,elfardy2013sentence}), morpheme (\citet{marcadet2005transformation,romsdorfer2007text,anand2014language}), words (\citet{vrehuuvrek2009language,adouane2017identification}), word combination (\citet{ccoltekin2016discriminating,singh2006study}), syllables (\citet{yeong2011applying,murthy2006language}), syntax (\citet{alex2005unsupervised,martinc2017pan}) and chunks (\citet{you2008identifying,elfardy2014aida}) as features. Even though LI task is considered a solved problem for distant languages (\citet{10.5555/1040196.1040208}), it is still challenging to identify closely related languages. Recently, more attention has been paid to similar or closely related languages, such as the Purvanchal languages we deal with in this work.

The more similar the languages are, the more challenging it is discriminate them or to identify them correctly for an LI system. Instead of n-gram based models, which did not consider sequences of characters or other units, the approaches to LI for closely related languages are mostly based on sequential algorithms. One of the earlier work on closely related Indian languages identification applied by pair-wise binary classification on Indo-Aryan (Hindi, Marathi, Punjabi, Oriya, Assamese and Bengali) and Dravidian (Tamil, Telugu, Kannada and Malayalam) language families. For the LI system, characters and aksharas are considered as the smallest appropriate unit to apply multiple learning methods (\citet{murthy2006language,singh2006study}). Recently VarDial 2018\footnote{\url{http://alt.qcri.org/vardial2018/index.php?id=campaign}} workshop included a shared task on language identification, which included Indo-Aryan languages, i.e., Hindi, Bhojpuri, Magahi, Braj and Awadhi. The dataset consisted of 15000 instances corresponding to each language. HeLI (\citet{jauhiainen2019language}) is a character level, n-gram language model with adaption technique, where the system provides 0.958\% macro F-score (\citet{zampieri2018language}).

We used a convolutional neural network (ConvNet) for the LI task at the character level. Earlier, ConvNet has been applied to text classification (\citet{zhang2015character}). The ConvNet takes the input of all possible characters used in the script of the language of the corpus (Devanagari for Bhojpuri, Maithili, Magahi and Hindi) to create embeddings of characters. Here, a total of 115 characters extracted from the corpus of all languages, written in Devanagari script include consonants, vowels, special symbols, digits etc. The input sentences are quantized lengthwise, which reduces the dimension of the character's one-hot vector. These vectors or embeddings are passed to the convolutional layer to extract language-specific features and select the optimal features by applying max-pooling over them. The pair of convolution and max-pooling layer is sequentially used six times. Finally, three fully connected layers with a non-linear function, followed by a pooling layer are applied, out of which the final fully connected layer defines the number of languages in our training corpus. 

\begin{table}[!h]
    \centering
    \caption{Language identification scores (\%) obtained by ConvNet}
    \begin{tabular}{cccc}
        \toprule
            \textbf{Language}    &      \textbf{Precision}    &   \textbf{Recall}  &  \textbf{F$_1$-score} \\
        \midrule
            Bhojpuri     &      98.94    &      99.67    &      99.30 \\
               Hindi     &      99.58    &      99.92    &      99.75 \\
              Magahi     &      99.59    &      98.36    &      98.97 \\
            Maithili     &      99.56    &      98.69    &      99.12 \\
        \hline
               \textbf{Macro avg.}     &      99.42   &       99.16   &       99.29 \\
            \textbf{Weighted avg.}     &      99.36    &      99.36   &       99.36 \\
        \bottomrule
    \end{tabular}
    \label{lang_ident_score}
\end{table}

The total instances in the dataset are 1,49,999, created by merging of Bhojpuri, Maithili, Magahi and Hindi monolingual corpus for performing LI using ConvNet. These instances correspond to sentences, of which 50149, 19463, 30520, and 49873 sentences are from Bhojpuri, Maithili, Magahi and Hindi, respectively. This dataset was divided into 80-20 ratio for training and validation of our model performance. The length of input sentences is fixed at 150 (with padding, as is usual while using deep neural networks) before training the model. 
The initial two pairs are of convolution and max-pooling using 256, 7 and 3 as convolution, kernel and pooling size. Remaining pairs have exploited 256, 3 and 3 (only on the last pair) as respective values. The units of fully-connected layers are 1024, 1024 and 4, which  are separated by a dropout layer (0.5 as dropout probability) to avoid overfitting of the model. The model was trained for 10 epochs with the minibatch size of 128 and Stochastic Gradient Descent (SGD) optimizer with the learning rate of 0.01.

The obtained results on the testing dataset (3604 instances, which are different from the training and the validation) data are given in Table \ref{lang_ident_score}. We have obtained the 99.36\% and 98.70\% accuracy on the testing and validation dataset, respectively, by this model.

While we get better language identification results for our experiments than the workshop on Indo-Aryan languages, this may be partly because we tried only with four languages.

Apart from providing a baseline for comparison for future work, this experiment on LI also acts as the last check on the quality of the data. These languages are quite close and almost none of the annotators use them for writing. Also, there is a complete lack of standardization and the boundaries between these languages are sometimes blurred as they are spoken mostly in a contiguous area. Each of them has numerous varieties, which change gradually at least according to the geographical location. It means that on the border areas, it is sometime not clear which language is being spoken, which is a common sociolinguistic phenomenon. Also, the annotators are from all over this area and they are familiar with different varieties of these languages, so there was often disagreement among them. The LI experiments at least shows that the data can be discriminated based on the language it is supposed to be from, using a standard LI algorithm for closely related languages.

\section{Conclusion and Future work}
We collected raw corpora from various sources (as mentioned in the Appendix) that cover different genres and domains for Bhojpuri, Magahi, and Maithili languages. After cleaning their raw corpora, we obtained 51374, 31251, and 19948 sentences, respectively. Out of these, 16067, 14669, and 12310 sentences are annotated for Part-of-speech tagging. We have performed statistical analysis at the character, word, morpheme, and syllable levels to exhibit the agglutinative and word-sharing properties. We have computed the unigram (or word), POS, and POS-word entropy of our corpus for the three languages and the n-gram character-based cross-entropy score among the languages, including Hindi, which show a degree of similarity among languages. Additionally, the manually annotated sentence reliability is estimated through Cohen's kappa Inter-Annotator Agreement and found to be 0.92, 0.64, and 0.74 for Bhojpuri, Magahi, and Maithili, respectively. The highest inter-annotator agreement score for Bhojpuri indicates that its annotation done by different linguists is found to be in the fairest agreement. We discussed some of the annotation issues for all the three languages. The POS tagged data was then annotated for Chunk tags: Bhojpuri (9695 sentences) and Maithili (1954 sentences). We built an initial POS tagger, Chunker and Language identifier tool by using these resources. Also, the bilingual dictionaries from these three Purvanchal languages to Hindi and synsets for IndoWordNet are prepared.

To the best of our knowledge, such corpora and corresponding syntactic annotation using POS tagging and Chunking, the synset (WordNet) and bilingual dictionary preparation is not available so far for these low-resource Purvanchal languages. The collected and annotated corpora can be used as the basis for the creation of further linguistic resources for these languages. These linguistic resources will then be equipped to create machine translation systems for these languages.

The results for the initial POS tagger and Chunker were reasonably good, i.e., good enough to be used as baselines for future work. The language identification results were better than reported earlier and they also confirm that the quality of the data, though problematic, is not completely useless. 

\section{Appendix}
\appendix
This section contains online sources from where different parts of the data for Purvanchal languages were collected and the example of their annotated sentences.
\\
\\
\textbf{Bhojpuri sources}\\
\url{http://www.bhojpuria.com/page=sahitya/index.html} \\ 
\url{http://www.thesundayindian.com/bh/} \\
\url{http://anjoria.com/} \\
\url{http://bhojpurika.com/} \\
\url{http://anjoria.com/v1/bhasha/bhasha.htm} \\
\url{http://www.readwhere.com/read/637950}
\\
\\
\textbf{Magahi sources}\\
\url{http://magahi-sahitya.blogspot.com/default}, the above blog taken contained a large fraction of Magahi colloquial (TeMTa) words.
\\
\\
\textbf{Maithili sources}\\
\url{http://www.esamaad.com/} \\
\url{http://www.maithilijindabaad.com/} \\
\url{https://www.bible.com/versions/1302-maint-jiv-at-n-s-at-ndesh} \\
\url{http://www.videha.co.in/} \\
\url{http://mithila-maithil-maithili-sites.blogspot.in/} \\
\url{http://maithili-katha.blogspot.in/} \\
\url{http://www.apnikahaani.com/2013/08/maithili-story-basaat.html} \\
\url{http://www.maithilijindabaad.com/?p=3902} \\
\url{https://hindipatal.wordpress.com/}
\\

\bibliographystyle{ACM-Reference-Format}
\bibliography{bmm_corpus.bib}

\appendix

\end{document}